\documentclass[10pt,twocolumn,letterpaper]{article}

\usepackage[pagenumbers]{cvpr} 
\usepackage{makecell}
\usepackage{tcolorbox}

\usepackage{graphicx}
\definecolor{cvprblue}{rgb}{0.21,0.49,0.74}
\usepackage[pagebackref,breaklinks,colorlinks,allcolors=cvprblue]{hyperref}
\usepackage{multirow}

\usepackage{algorithm}
\usepackage{algpseudocode}
\usepackage{array,booktabs}
\usepackage{tcolorbox}
\usepackage{tabularx}
\tcbuselibrary{skins}
\usepackage{hyperref}
\definecolor{ToolGray}{RGB}{80, 80, 80} 

\newtcolorbox{ToolCard}[2][]{
  colback=white, 
  colframe=ToolGray,       
  colbacktitle=ToolGray,   
  coltitle=white,          
  fonttitle=\bfseries\large, 
  title={#2},
  arc=1mm,                 
  boxrule=0.5pt,           
  left=2mm, right=2mm, top=2mm, bottom=2mm, 
  enhanced,                
  #1                       
}

\newcommand{\ToolKey}[1]{\par\noindent\textbf{#1}\hspace{0.5em}}
\def\paperID{} 
\def\confName{CVPR}
\def\confYear{2026}

\title{MANSION: Multi-floor lANguage-to-3D Scene generatIOn for loNg-horizon tasks}

\author{
Lirong Che$^{*,1,2}$ \quad
Shuo Wen$^{*,3}_{\S}$ \quad
Shan Huang$^{1}$ \quad
Chuang Wang$^{2}$ \\
Yuzhe Yang$^{2}$ \quad
Gregory Dudek$^{3}$ \quad
Xueqian Wang$^{\dagger,1}$ \quad
Jian Su$^{\dagger,2}$ \\
\fontsize{9pt}{11pt}\selectfont 
$^{1}$Tsinghua University \quad
$^{2}$AgiBot \quad
$^{3}$McGill University, MILA - Quebec AI Institute
}

\begin{document}

\twocolumn[{
  \maketitle
  \vspace{-0.8em}
  \begin{center}
    \includegraphics[width=\linewidth]{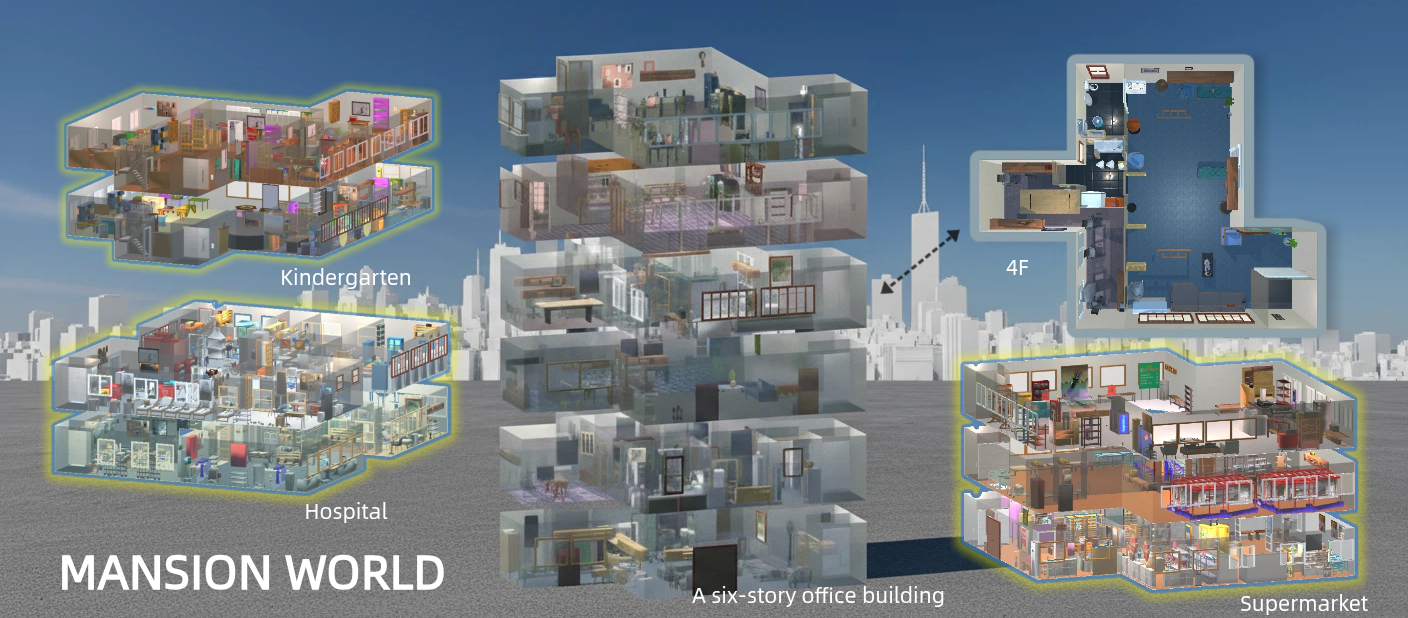}
    \captionof{figure}{
        \textbf{MansionWorld: The first building-scale dataset for long-horizon embodied AI tasks.} 
        Generated by our \textbf{MANSION} framework, this dataset represents the first large-scale collection of multi-story, customizable themed environments. The visualization highlights four representative examples: Kindergarten, Hospital, Supermarket, and a Six-story Office Building, which feature complex functional zoning and fully navigable vertical connections to support long-horizon, cross-floor embodied AI tasks. You can access the MansionWorld dataset at: {\footnotesize\href{https://huggingface.co/datasets/superbigsaw/MansionWorld}{Link to MansionWorld}}
    }
    \label{fig:teaser}
  \end{center}
  \vspace{0.8em}
}]
\begingroup
\renewcommand\thefootnote{}
\footnotetext{* Equal contribution. $\dagger$ Corresponding authors. $\S$  Work done during an internship at Agibot}
\endgroup

\begin{abstract}
Real-world robotic tasks are long-horizon and often span multiple floors, demanding rich spatial reasoning. However, existing embodied benchmarks are largely confined to single-floor in-house environments, failing to reflect the complexity of real-world tasks. We introduce \textbf{MANSION}, the first language-driven framework for generating building-scale, multi-floor 3D environments. Being aware of vertical structural constraints, MANSION generates realistic, navigable whole-building structures with diverse, human-friendly scenes, enabling the development and evaluation of cross-floor long-horizon tasks. Building on this framework, we release \textbf{MansionWorld}, a dataset of over 1,000 diverse buildings ranging from hospitals to offices, alongside a Task-Semantic Scene Editing Agent that customizes these environments using open-vocabulary commands to meet specific user needs. Benchmarking reveals that state-of-the-art agents degrade sharply in our settings, establishing MANSION as a critical testbed for the next generation of spatial reasoning and planning.
\end{abstract}
    
\section{Introduction}
\label{sec:intro}

The ultimate goal of Embodied AI is to build agents that can autonomously reason and accomplish any difficult tasks in the complex real world. Many critical applications, ranging from package delivery in offices, supply transport in hospitals, to multi-step chores at home, are inherently long-horizon and at the building scale. These tasks demand not only low-level skills such as navigation and object manipulation~\cite{brohan2022rt,zitkovich2023rt,driess2023palme}, but also strong capabilities for long-horizon spatial planning, reasoning, and memory~\cite{huang2022inner,singh2022progprompt,anwar2025remembr}. Recent work has underscored the need for unified planning of manipulation and navigation in constrained, building-wide settings~\cite{shah2025bumble,li2023behavior1k}, yet, to the best of our knowledge, no current benchmarks match this level of complexity. 

One central challenge is the mismatch between the limited existing scene resources and the growing demand from embodied AI and 3D scene generation algorithms for large-scale, diverse, and interactive simulation environments. Although real-world scanned datasets provide high-fidelity geometry and textures~\cite{Chang2017Matterport3D,Baruch2021_ARKitScenes}, the data is expensive to collect and hard to recycle for downstream editing or reconfiguration, making it difficult to match task requirements. Synthetic environments, generated either procedurally ~\cite{Deitke2022_ProcTHOR} or by data-driven, LLM-based approaches~\cite{shabani2023housediffusion,Yang_2024_Holodeck}, largely focus on single-floor rooms or apartment-scale layouts, and rarely model vertical structure, inter-floor portals, or transit facilities such as elevators and staircases explicitly. As a result, the absence of scalable, easily reconfigurable, and building-scale simulation environments has become a key bottleneck for progress in embodied AI, directly limiting research on long-horizon embodied tasks with a focus on spatial reasoning.


    

    

To address these problems, we introduce \emph{MANSION}, a language-driven framework for building-scale environment generation and long-horizon task evaluation. Building on top of it, we also introduce the generated dataset and embodied evaluation ecosystem, \emph{MansionWorld}. See Fig.~\ref{fig:teaser}. Our webpage containing the code can be found at: \href{https://agibotgeneral.github.io/mansion-site/}{Mansion Webpage}. We summarize our main contributions as follows: 

\begin{itemize}
    \item We propose a hybrid multimodal large language model (MLLM)–geometry pipeline that turns natural-language instructions into complete multi-story buildings in 3D scenes, represented as semantically grounded, vertically aligned vector floorplans, with innovative spatial constraints. These layouts can be used off-the-shelf in AI2-THOR~\cite{kolve2017ai2} and exported to other physics simulators.
    
    \item We extend the original AI2-THOR~\cite{kolve2017ai2} framework with reusable tunneling assets and cross-floor skill APIs, enabling building-scale, multi-floor embodied tasks to be defined and evaluated.
    
    \item We design a Task-Semantic Scene Editing Agent that transforms the generated static buildings into an adaptable playground by enabling fine-grained scene modifications, allowing the versatile recycling of the same environment to meet the needs of a variety of tasks.
    
    \item We release \emph{MansionWorld}, a large-scale ecosystem of over \textbf{1,000} diverse, interactive multi-floor buildings spanning residential, office, and public facility domains. Through experiments on floorplan generation and embodied benchmarks, we show that our constrained-growth solver generalizes beyond standard residential datasets, while state-of-the-art embodied agents exhibit sharp degradation on our multi-floor tasks, underscoring the difficulty and value of this setting.
\end{itemize}

\section{Related Work}
\label{sec:related_work}

\textbf{Long-Horizon and Multi-Level Embodied Tasks.}
Driven by recent progress in embodied manipulation and navigation~\cite{driess2023palme, brohan2022rt, zitkovich2023rt, kim24openvla, brohan2023can, shah2023lm, huang2022inner, liang2023code, singh2022progprompt, huang2023voxposer, rana2023sayplan}, robotic systems are increasingly tackling complex, long-horizon tasks at the building scale. To support these extended operations, several approaches incorporate spatio-temporal memory and topological representations to maintain awareness of the environment state~\cite{anwar2025remembr, lei2025stma, zheng2024esceme}. However, existing benchmarks for embodied tasks remain oversimplified, with isolated focuses on either local close-range manipulation or navigation under basic spatial connectivity, which lack the modeling of interactions with architectural elements such as doors and elevators~\cite{anderson2018vln, ku2020roomacrossroom, krantz2020beyond, Shridhar_2020_CVPR, padmakumar2022teach}. As a result, such benchmarks failed to present the key challenges in real multi-story buildings, including cross-floor mobility, structural interactions, and the joint demands of long-horizon planning and memory. This highlights the need for executable multi-floor environment benchmarks that can systematically evaluate navigation, interaction, planning, and memory in a unified setting~\cite{shah2025bumble, li2023behavior1k}.

\textbf{Floorplan Generation.}
As a foundational component for structured scene synthesis, the earliest methods in the field of floorplan used finite state grammars~\cite{rao2007randomized} or L-systems~\cite{antoniuk2018system,goel1991some}.
More recent methods utilized graph neural networks to convert room adjacencies into layouts~\cite{hu2020graph2plan}, while recent diffusion models~\cite{shabani2023housediffusion, hu2025gsdiff} and LLM-guided paradigms~\cite{qin2024chathousediffusion, zong2024housellm} have enabled the direct generation of diverse vector floorplans from text.
Despite their impressive performance on the topological correctness and diversity of single-story residences, these methods are almost universally confined to single-story layouts. 
They neither model the alignment of exterior contours between floors nor enforce the spatial consistency of vertical cores like stairs and elevator shafts. 
Furthermore, their output is typically a static vector or raster image, which lacks the executable semantics required for direct use in simulation and task planning, making them unsuitable for cross-floor tasks.

\textbf{Language-driven 3D Scene Generation. }To overcome the limitations of manual or scanned datasets~\cite{Chang2017Matterport3D, szot2021habitat2, Baruch2021_ARKitScenes, wang2024grutopia} and procedural generation~\cite{Deitke2022_ProcTHOR} in scalability and semantics, recent research has leveraged LLMs as ``scene directors" to achieve controllable 3D synthesis. Methods like Holodeck generate layouts based on spatial constraints to support downstream tasks~\cite{Yang_2024_Holodeck, Deitke2023_ObjaverseXL}; SceneCraft emphasizes cross-room visual consistency~\cite{Yang2024SceneCraft}; and SceneWeaver enhances physical plausibility through reflection cycles~\cite{Yang2025_SceneWeaver}. Despite progress in visual fidelity and \emph{in-plane} task support, these methods remain universally confined to single-story layouts. They fail to model or validate cross-floor connectivity via vertical core structures. Consequently, their topologically ``flat" environments lack the complexity for long-horizon, cross-floor planning, hindering scaling to realistic, building-scale tasks. In contrast, our work generates building-scale environments that treat vertical structures as an explicit constraint, ensuring provable navigability and task-readiness.

\begin{table}[t]
\centering
\caption{Comparison of floorplan generation methods. \textbf{Bdry.}/\textbf{Topo.}/\textbf{Vert.} denote Boundary/Topology/Vertical structure. Boundary/Topology: controllable conditioning at test time. Vertical structure indicates cross-floor aligned cores (walls/rooms/regions) that persist across floors. Room type: resident vs. open-vocab.}
\label{tab:method_comparison}
\footnotesize
\setlength{\tabcolsep}{2pt}
\renewcommand{\arraystretch}{0.9}
\resizebox{\columnwidth}{!}{
\begin{tabular}{>{\raggedright\arraybackslash}p{2.9cm} c c c c >{\raggedright\arraybackslash}p{1.8cm}}
\toprule
Method & Type & Bdry. & Topo. & Vert. core & Room type \\
\midrule
Graph2Plan~\cite{hu2020graph2plan}
  & model-based
  & \checkmark
  & \checkmark
  & $\times$
  & resident \\

HouseDiffusion~\cite{shabani2023housediffusion}
  & model-based
  & $\times$
  & \checkmark
  & $\times$
  & resident \\

GSDiff~\cite{hu2025gsdiff}
  & model-based
  & \checkmark
  & $\times$
  & $\times$
  & resident \\
\midrule
ProcTHOR~\cite{Deitke2022_ProcTHOR}
  & rule-based
  & \checkmark
  & $\times$
  & $\times$
  & resident \\

Holodeck~\cite{Yang_2024_Holodeck}
  & LLM
  & $\times$
  & $\times$
  & $\times$
  & \textbf{open-vocab} \\
\midrule
AnyHome~\cite{fu2024anyhome} 
  & LLM+model
  & $\times$
  & \checkmark
  & $\times$
  & resident \\

ChatHouseDiffusion \cite{qin2024chathousediffusion}
  & LLM+model
  & \checkmark
  & \checkmark
  & $\times$
  & resident \\

HouseLLM~\cite{zong2024housellm}
  & LLM+model
  & $\times$
  & \checkmark
  & $\times$
  & resident \\
\midrule
\textbf{MANSION}
  & \textbf{LLM+rules}
  & \textbf{\checkmark}
  & \textbf{\checkmark}
  & \textbf{\checkmark}
  & \textbf{open-vocab} \\
\bottomrule
\end{tabular}
}
\end{table}
\section{MANSION}

While effective for single-floor, existing floorplan generators fail to scale to multi-story buildings due to two fundamental limitations. First, these generators lack vertical consistency, making them unable to align exterior contours or critical vertical cores across floors. Secondly, their data-driven nature restricts them to `closed-world' residential datasets( see Table \ref{tab:method_comparison}), failing to generalize to out-of-distribution types.

MANSION systematically solves these issues. Our framework uniquely enforces vertical alignment as a first-class hard constraint, ensuring 3D structural validity. We also employ an MLLM-driven hybrid architecture that decouples high-level semantics from low-level geometry. This design achieves true open-world scalability, generating diverse building types without new data or retraining.
\begin{figure*}[t]
    \centering
    \includegraphics[width=\textwidth]{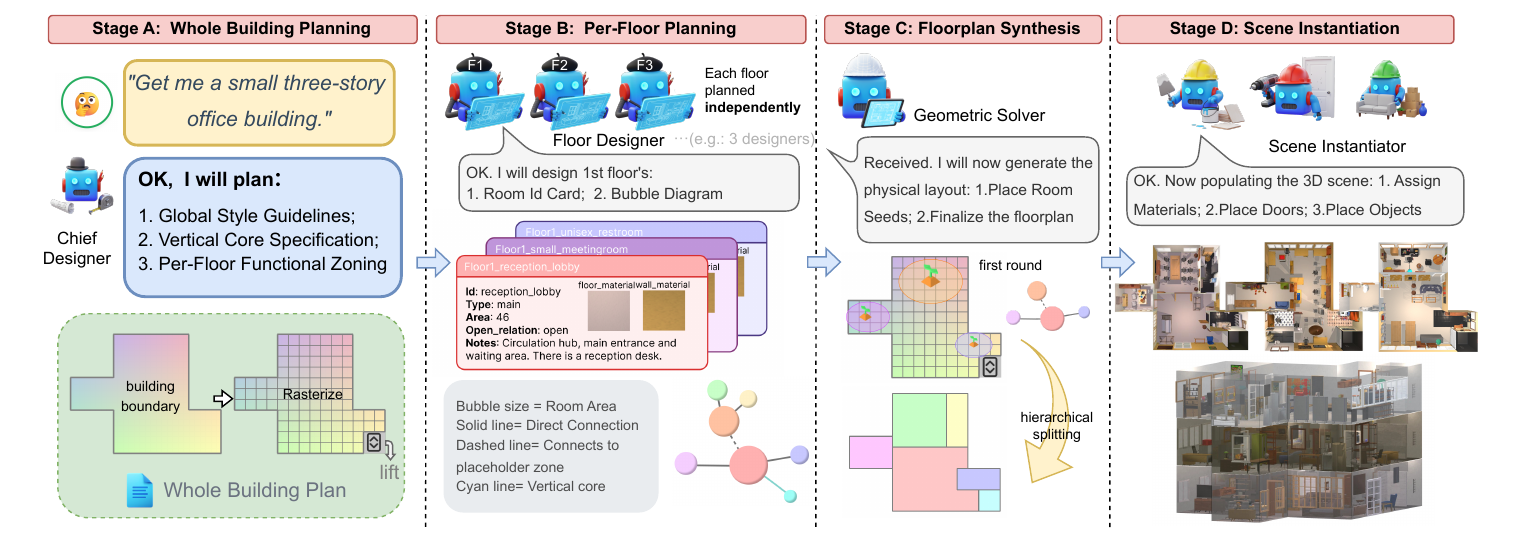}
    \caption{
        Overview of the MANSION framework: a multi-agent-driven pipeline for generating multi-story 3D buildings from natural language.
        The process includes: (A) Whole Building Planning, (B) Per-Floor Planning, (C) Floorplan Synthesis, and (D) Scene Instantiation.
    }
    \label{fig:mansion_pipeline}
\end{figure*}
\subsection{MANSION Framework}
\label{sec:framework}
MANSION is a hierarchical multi-agent framework, as illustrated in Fig.~\ref{fig:mansion_pipeline}, that progressively transforms natural-language-specified building requirements into interactive multi-floor 3D scenes. Throughout the generation pipeline, floorplan generation is the key bridge between high-level semantic planning and downstream scene instantiation. We therefore first formalize it as a verifiable constrained search problem, and then present the scene instantiation process.

\textbf{Floorplan generation.}
We begin by formalizing the generation task. Let the outer footprint of each floor be an orthogonal polygon $P_f$ (where $f$ indexes floors), and let $\mathcal{V}$ denote the set of vertical structures (stairs, elevators, shafts, etc.). We denote by $Q_{f,v} \subseteq P_f$ the geometric footprint of vertical core $v \in \mathcal{V}$ on floor $f$, and only plan rooms in the free region
\[
    \Omega_f = P_f \setminus \bigcup_{v \in \mathcal{V}} Q_{f,v}.
\]
The high-level layout specification is given as a bubble diagram $\mathcal{G} = (\mathcal{R}, \mathcal{E})$~\cite{hu2020graph2plan,shabani2023housediffusion,hu2025gsdiff}. Each node $r \in \mathcal{R}$ corresponds to a room (or semantic region) to be instantiated, with target area $a_r$; an edge $(r_i, r_j) \in \mathcal{E}$ indicates that an adjacency or connectivity relation should exist between $r_i$ and $r_j$ in the final layout, and may include room--room, room--vertical-core, and cross-floor relations.

We formulate floorplan synthesis as a \emph{verifiable search over a candidate set}:
\[\label{search}
    L^\star
    =
    \arg\max_{L \in \mathcal{C}} \mathrm{Score}(L; \mathbf{w})
    \quad \text{s.t.} \quad
    \mathrm{Topo}(L, \mathcal{G}) = \mathrm{true},\]

where $L$ is the room partition on the current floor (or the entire building), represented as a set of polygonal regions inside $\Omega_f$, and $\mathcal{C}$ is a discrete candidate set produced by sampling and constrained growth. The function $\mathrm{Score}$ is an energy-based objective used to rank feasible candidates within $\mathcal{C}$. 

To solve the above search problem, we organize floorplan generation as a multi-MLLM subsystem orchestrated by LangGraph. The core idea is not to let the MLLM directly regress complete room polygons, but to first decompose high-level semantic requirements into an intermediate representation that is more compatible with current MLLM capabilities, and then perform verifiable search under these intermediate constraints using a geometric solver.

Specifically, a \emph{building-level planning node} first determines cross-floor functional zones, target area allocation, and global stylistic preferences from the user's natural-language specification and the building footprint, thereby ensuring semantic and visual consistency across the whole building. These global constraints are then dispatched to per-floor \emph{floor-planning nodes}, each of which generates a bubble diagram $\mathcal{G}_f = (\mathcal{R}_f,\mathcal{E}_f)$ on the corresponding free region $\Omega_f$, specifying the room set, target area $a_r$, and adjacency relations to vertical cores and other rooms.

Before geometric solving, we rasterize each $\Omega_f$ into a 2D grid and pass it to a dedicated \emph{cutting MLLM node}. This node provides an initial growth seed $c_r \in \Omega_f$ for each target room, offering coarse spatial guidance for room placement. Prior work suggests that modern MLLMs have significantly improved visual pointing and spatial grounding capabilities, making such a grid-based seed proposal interface feasible in medium-scale scenes~\cite{park2025rvlm,neo2025coordinate}.

To avoid the high combinatorial complexity of deciding all room locations at once, we further adopt a \emph{hierarchical splitting} strategy. Starting from a circulation hub node in the bubble diagram, the cutting MLLM only needs to select one valid child room from the current topological front at each step and provide its local seed within the parent region.

The solver then takes this seed together with the target area as priors, and realizes the local split using our \emph{single-cut} solver, a topology-aware variant of Lopes-style constrained growth~\cite{lopes2010constrained}. It generates local candidate partitions inside the parent region, filters out candidates that violate already-realized topological relations, and ranks the remaining ones with an interpretable energy function, accepting the highest-scoring partition. This process iterates along the topological front until all room nodes on the current floor have been partitioned.

\textbf{Scene instantiation.}
After obtaining the room partition on each floor, we instantiate the layout into interactive AI2-THOR scenes~\cite{kolve2017ai2}, including architectural elements, doors, and objects. 

Our instantiation follows a two-level, progressive planning design. First, a building-level ``chief designer'' node determines the global visual style once at the beginning (e.g., material palette and color scheme), ensuring cross-floor consistency. Then, as each floor-planning node generates its bubble diagram, it attaches a room card to each room node, encoding material preferences, openness type, and finer-grained functional requirements. Downstream instantiation nodes (material assignment, opening and door generation, and object placement) realize these room cards under already-satisfied topological constraints, so the final scene remains consistent with the high-level design in both visual style and connectivity.

We follow the LLM+rule-based placement paradigm of HOLODECK~\cite{Yang_2024_Holodeck}, but shift the design philosophy from quantity-first to usability- and quality-first. First, we enforce hard reachability as a non-negotiable constraint: only objects with sufficient surrounding clearance that the robot can navigate to are retained. Second, to prevent object clustering in large rooms, we introduce anchor-based groups, where an anchor object carries a global spatial tag (\texttt{edge}/\texttt{middle}) and remaining group members are solved in the anchor's local reference frame, yielding more uniform spatial distribution and fewer placement conflicts. Third, we add two structured relation primitives, \texttt{matrix} and \texttt{paired}, for grid-pattern and symmetric co-placement, respectively, enabling orderly arrangement of desks, shelves, and chairs in non-residential environments such as classrooms, libraries, and open-plan offices. Finally, we adopt a priority-aware placement order combined with quality-first pruning: wall-adjacent and structured-pattern objects are placed first to minimize interference with navigation corridors; any candidate that violates reachability or falls below quality thresholds is discarded outright rather than retained via soft-constraint relaxation.

The full generation pipeline detail can be found in Supp.~\ref{sec:appendix-pipeline}, and the object placement algorithmic details can be found in Supp.~\ref{sec:obj_placement}. 

\subsection{MANSION Ecosystem}
The MANSION Ecosystem is built on top of the generation environments for complex tasks. It comprises three key components: a large-scale dataset of 1,000 buildings, new stair and elevator assets with enhanced agent cross-floor navigation capabilities, and a Task-Semantic Scene Editing Agent for defining limitless embodied tasks within the scenes.

\begin{figure}[h]
    \centering
    \includegraphics[width=1\columnwidth]{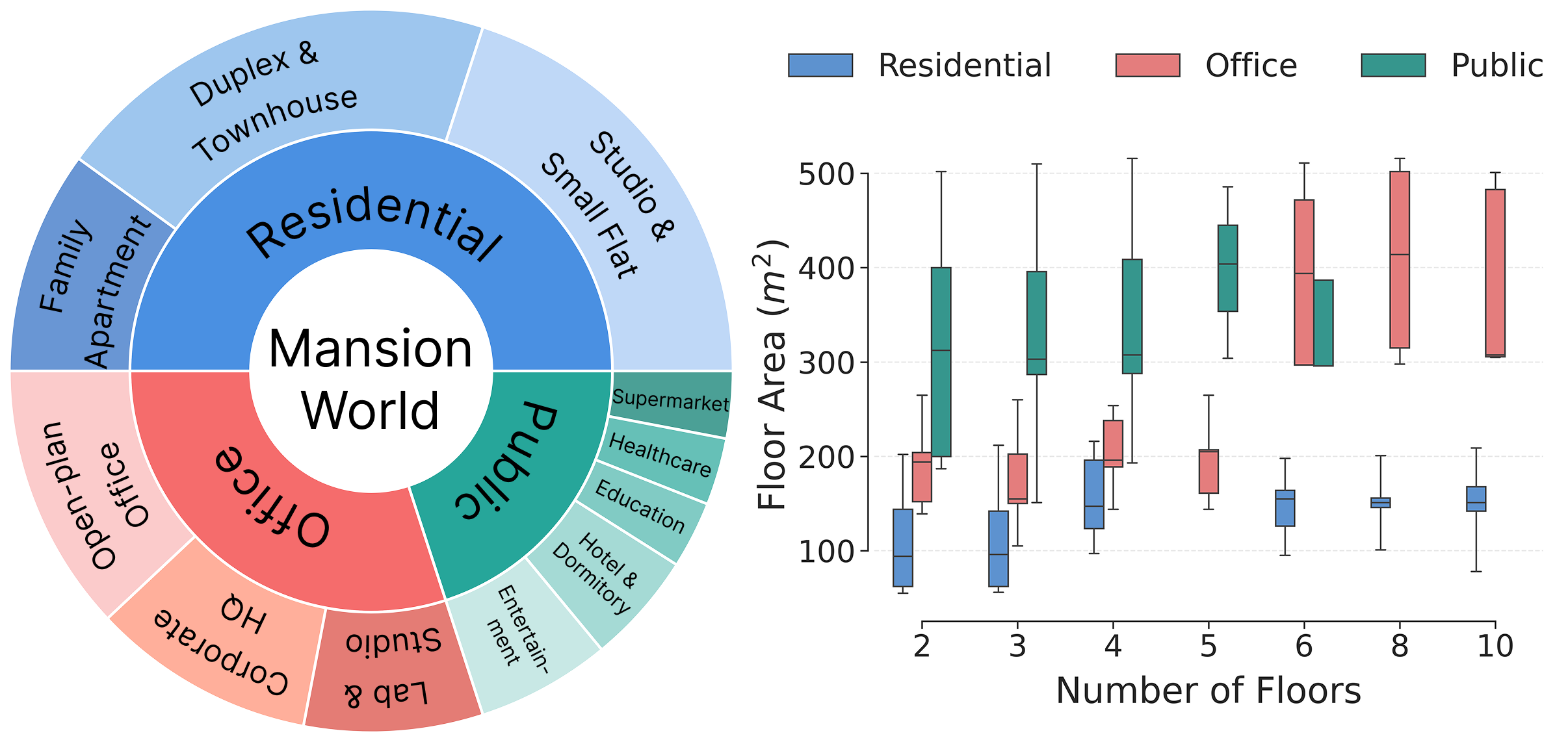}
    \caption{MansionWorld statistics: functional composition and floor-area distributions across different floor counts. }
    \label{fig:dataset}
\end{figure}

\textbf{MansionWorld: A Large-Scale Building-Scale Dataset.} Based on the MANSION generation pipeline introduced in the previous section, we build and release \textbf{MansionWorld}: a new, large-scale dataset of diverse, interactive multi-story buildings. Moving beyond existing residential-based benchmarks, MansionWorld provides unprecedented diversity in building types, covering functional, non-residential environments such as office buildings, hospitals, schools, supermarkets, and entertainment centers. As shown in Fig.~\ref{fig:dataset}, MansionWorld spans diverse functional categories and physical scales. The dataset features over \textbf{1,000} unique buildings, with structures ranging from 2 to 10 stories in height and totaling over \textbf{10,000} individual rooms. To support the broadest possible community research, we also provide tools to export the scene geometry and semantics to other popular platforms, such as Blender for high-fidelity rendering and NVIDIA Isaac Sim for physics-based simulation.

\begin{figure*}[t]
  \centering
  \includegraphics[width=\textwidth]{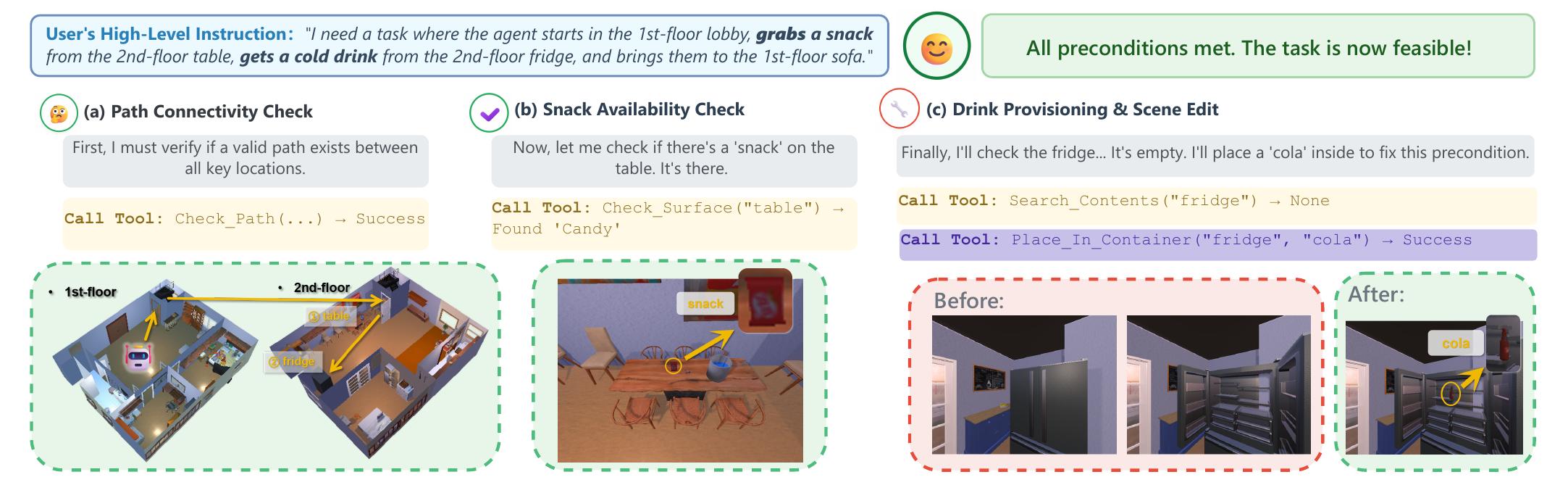} 
  \caption{The ``Check-and-Provision" workflow of our Task-Semantic Scene Editing Agent. The agent first decomposes a high-level instruction (``bring a snack and a drink to the sofa'') into preconditions. It then sequentially performs a (a) Path Connectivity Check, an (b) Object Availability Check, and an (c) Object Provisioning \& Scene Edit to ensure the task is executable before generation.}
  \label{fig:check}
\end{figure*}

\textbf{Cross-Floor Mobility via Stairs and Elevators.} To enable MansionWorld for the complex, building-scale tasks it is designed for, we extend the core capabilities of the AI2-THOR~\cite{kolve2017ai2} simulator. We design and integrate two crucial categories of interactive assets: multi-flight stairwells (Stairs) and functional elevators (Elevators). Beyond the assets themselves, we develop a suite of high-level atomic skill APIs (e.g., \texttt{UseStairs}, \texttt{CallElevator}, \texttt{UseElevator}) that encapsulate the interaction logic. These APIs are critical as they handle the underlying scene-to-scene transition management; for instance, executing \texttt{UseStairs} seamlessly unloads the current scene graph and loads the target floor, placing the agent at the correct landing. This, for the first time on the platform, provides agents with robust and seamless cross-floor navigation, a fundamental prerequisite for any building-scale task.

\textbf{Task-Semantic Scene Editing Agent.} Once static multi-floor buildings are generated, a core challenge is to make them versatile enough to efficiently support diverse embodied AI tasks. Generating a new environment for each individual task is not only inefficient, but hard-coding task requirements into the design process also over-constrains the layout, making it less useful for new tasks subsequently.

To address this problem, we propose a Task-Semantic Scene Editing Agent. It is driven by an MLLM controller that understands high-level natural language instructions and modifies the scene through a series of controlled tool calls to satisfy task preconditions. The agent's core capability lies in translating a user's high-level task directive into a sequence of scene edits that ensure task executability. Rather than allowing the MLLM to directly edit raw scene data, we provide it with a small yet expressive set of tool APIs encapsulated on top of AI2-THOR. These tools permit the agent to query scene structure, retrieve assets, and perform object and container manipulations.

As illustrated in Fig.~\ref{fig:check}, when a user provides a complex, multi-floor task instruction such as, ``I need a task where the agent starts in the 1st-floor lobby, grabs a snack from the 2nd-floor table, gets a cold drink from the 2nd-floor fridge, and brings them to the 1st-floor sofa," the agent does not proceed to execution immediately. Instead, it first decomposes the task into a set of necessary preconditions and initiates a ``Check-and-Provision" workflow. Through this ``think-verify-act" loop, all preconditions are met, rendering a task that was initially infeasible due to missing objects fully executable. Furthermore, these edits can be persisted, allowing for the creation and reuse of multiple task variations.

This editing approach is designed to complement, rather than replace, existing text-to-environment generation techniques. Such systems typically focus on synthesizing new 3D environments from scratch, often prioritizing visual diversity. In contrast, we assume a pre-generated, structurally stable corpus of buildings and apply minimal, task-oriented edits to specialize them for specific embodied tasks, preserving structural realism while crucially ensuring executability.

The core advantage of this design is the dramatic enhancement of reusability. A single building can dynamically host a vast number of language-defined, reproducible tasks. This effectively transforms our building dataset into a task-semantic playground for studying long-horizon, compositional embodied agents, all without the need to regenerate an entire environment for each new task.

\section{Experiments}
\subsection{Floorplan Generation Algorithm}
\label{sec:floor_gen}
We evaluate our method on the T2D dataset~\cite{leng2023tell2design} by applying a unified vectorized pre- and post-processing pipeline to the underlying geometry, in order to keep the evaluation protocol comparable to prior work. T2D is essentially a post-processed derivative of RPLAN~\cite{wu2019data}, so we directly read each room's polygonal contour and vertex coordinates from the original JSON annotations and obtain a vector floor plan in the world-coordinate space. We then uniformly scale each floor plan onto a fixed-resolution grid and rasterize it into a room-level semantic label map after rounding the scaled vertex coordinates to the nearest integer grid points. In this raster space, we run our MLLM-based point selection and hierarchical growth algorithm to generate the corresponding predicted label maps. At evaluation time, we compare prediction and ground-truth masks at the same resolution and report pixel-level $\text{micro-IoU}$ (overall IoU over all pixels) and $\text{macro-IoU}$ (class-averaged IoU over room categories). Compared to the official T2D implementation, we adopt a ``polygon-to-raster mask'' pipeline instead of the original interface; however, ground truth and predictions share exactly the same scaling and rasterization process, and the grid resolution is sufficiently high relative to the original integer coordinates, so the additional quantization error is negligible. The resulting IoU measurements are therefore theoretically equivalent to the original definition in T2D~\cite{leng2023tell2design}, up to minor finite-resolution effects.

We follow the experimental protocol of ChatHouseDiffusion (CHD)~\cite{qin2024chathousediffusion} and organize the comparison into two main parts. First, to disentangle the effect of large language models (LLMs), we directly compare CHD's core diffusion model with our constrained growth algorithm under the manual annotation (MA) setting. In CHD, MA refers to using JSON data extracted directly from the floor plans as geometric supervision. For a fair comparison, we adopt the same setting: we extract room centroids from the original annotations as seed positions and use the ground-truth room areas as inputs to our constrained growth module.

\begin{table}[t]
    \centering
    \caption{IoU scores under different configurations on T2D}
    \label{tab:t2d_iou}
    \begin{tabular}{lcc}
        \toprule
        \textbf{Method} & \textbf{Micro-IoU} & \textbf{Macro-IoU} \\
        \midrule
        Obj-GAN~\cite{li2019object} & 10.68 & 8.44 \\
        CogView~\cite{ding2021cogview} & 13.30 & 11.43 \\
        Imagen~\cite{saharia2022photorealistic} & 12.17 & 14.96 \\
        T2D & 54.34 & 53.30 \\
        \midrule
        CHD (moonshot)         & 60.09 & 56.09 \\
        \textbf{CHD (gemini-2.5-pro)}  & \textbf{76.34} & \textbf{72.24} \\
        \textbf{CHD (MA)}              & \textbf{82.81} & \textbf{79.04} \\
        Ours (moonshot)                      & 42.33 & 40.95 \\
        \textbf{Ours (gemini-2.5-pro)}                & \textbf{69.98} & \textbf{66.40} \\
        \textbf{Ours (MA)}                            & \textbf{81.67} & \textbf{80.66} \\
        \bottomrule
    \end{tabular}
\end{table}

As shown in Table~\ref{tab:t2d_iou}, on the T2D dataset our method (Ours-MA) achieves performance comparable to CHD-MA under the same MA setting. This result provides strong evidence that the proposed constrained growth algorithm can effectively fit the complex room layouts commonly seen in residential environments. We then evaluate the end-to-end pipeline and compare different MLLMs, including Moonshot-v1-8k (used in the original CHD paper) and Gemini-2.5-Pro. When using the earlier Moonshot model, our method lags significantly behind CHD. However, when both methods are driven by the stronger Gemini-2.5-Pro, the performance gap narrows substantially. This observation is consistent with our design intuition: our method fully delegates semantic understanding and spatial pointing to the LLM, while the constrained growth module focuses solely on geometric solving. As the LLM's pointing capability (i.e., spatial localization accuracy) improves, the quality of the predicted seeds and area priors improves accordingly, leading to more accurate overall layouts.

To further assess the generalization ability of our method in more complex and realistic settings, we conduct experiments on a 1K-sample subset of the ResPlan dataset~\cite{abouagour2025resplan}. ResPlan provides native vector polygons and room-level topology, and compared with the residential scenes in T2D/RPLAN, it exhibits substantially larger room counts and richer structural complexity. In our sampled subset, nearly $50\%$ of the floor plans contain more than eight rooms, whereas the RPLAN-based training setup of CHD is limited to at most eight rooms. For a fair comparison, we map all room types in ResPlan to the standard category space used by CHD.

As summarized in Table~\ref{tab:resplan_iou}, although CHD performs strongly on T2D, its performance on the more challenging ResPlan-1K benchmark is unsatisfactory (we do not retrain CHD on ResPlan-1K, but evaluate it in a zero-shot setting, emphasizing its ability to generalize from RPLAN/T2D to more complex, non-residential layouts). Even when we restrict evaluation to the subset with at most eight rooms (CHD 8minus-MA), the IoU remains extremely low. We hypothesize that this degradation is related to the observation in MSD~\cite{vanengelenburg2024msd} that the RPLAN dataset ``contains a serious amount of near-duplicates,'' which may limit the diversity of CHD's training distribution and harm its generalization to more realistic and structurally diverse layouts in ResPlan. In contrast, our method achieves a micro-IoU of $76.74\%$ under the MA setting on ResPlan-1K, demonstrating that the constrained growth algorithm maintains strong layout-fitting ability even in complex scenarios.
\begin{table}[t]
    \centering
    \caption{IoU scores under different configurations on Resplan-1k}
    \label{tab:resplan_iou}
    \begin{tabular}{lcc}
        \toprule
        \textbf{Method} & \textbf{Micro-IoU} & \textbf{Macro-IoU} \\
        \midrule
        CHD (gemini-2.5-pro)                          & 29.36 & 22.25 \\
        CHD (8minus-MA)                               & 36.12  & 26.14  \\
        CHD (MA)                                      & 33.49 & 25.39 \\
        \makecell[l]{Ours (gemini-2.5-pro) \\ \quad w/o hierarchical splitting} & 45.65 & 42.42 \\
        \textbf{Ours (gemini-2.5-pro)}                & \textbf{63.56} & \textbf{61.65} \\
        \textbf{Ours (MA)}                            & \textbf{76.74} & \textbf{76.64} \\
        \bottomrule
    \end{tabular}
\end{table}
We further perform an ablation study to validate the importance of our proposed iterative splitting strategy. Specifically, we compare the full method, which uses iterative cutting, against a variant that requires the MLLM to output all room seed coordinates in a single step. We observe a significant drop in micro-IoU in the one-shot variant. Our analysis suggests that while the MLLM provides relatively stable area priors, one-shot prediction of all initial seed positions suffers from large pointing errors. This result highlights the importance of the iterative splitting strategy in reducing task complexity and improving spatial localization accuracy.

\subsection{Object Placement Evaluation}
\label{sec:object_placement_eval}
To evaluate the performance of our object placement module, we conduct comparative experiments on four room types, covering both residential and non-residential environments, as well as regular and irregular room geometries. We compare our method with representative open-vocabulary 3D scene synthesis approaches, including LayoutGPT~\citep{feng2023layoutgpt} and Holodeck~\citep{Yang_2024_Holodeck}. For each room type and each method, we perform 10 independent runs for evaluation. Our evaluation protocol generally follows SceneWeaver~\citep{Yang2025_SceneWeaver}, while being adapted to embodied-task requirements. In particular, we introduce an additional reachability metric to measure whether target objects in the generated scene can be effectively approached and interacted with by the robot.

We further conduct a user study with 52 participants. The details of this user study can be found in Supp.~\ref{sec:user_study}. Note that we do not directly compare with SceneWeaver~\citep{Yang2025_SceneWeaver}. This is because our current implementation is still a one-shot placement module without reflection-based iterative optimization. Therefore, the purpose of this experiment is to validate the effectiveness of the module as a one-shot placement solver, and to show its potential as a foundation for future iterative refinement. 

\begin{table*}[t]
\centering
\normalsize
\setlength{\tabcolsep}{1.8pt}
\renewcommand{\arraystretch}{0.92}
\caption{\textbf{object placement quantitative comparison.}
We report the average number of placed objects (\#Obj, with small items in parentheses), out-of-boundary objects (\#OB), Layout-level collided object pairs (\#CN), floor-object reachability (\#Rch, \%), and user-study preference scores (\%) for Realism (Real.), Diversity (Div.), and Layout (Lay.).}
\label{tab:single_room_object}
\begin{tabular*}{\textwidth}{@{\extracolsep{\fill}}lccccccc ccccccc@{}}
\toprule
\multirow{2}{*}{Method} &
\multicolumn{7}{c}{Bedroom (4$\times$4 m, rect.)} &
\multicolumn{7}{c}{Classroom (8$\times$8 m, rect.)} \\
\cmidrule(lr){2-8}\cmidrule(l){9-15}
& {\small \#Obj$\uparrow$} & {\small \#OB$\downarrow$} & {\small \#CN$\downarrow$} & {\small \#Rch$\uparrow$} & {\small Real.$\uparrow$} & {\small Div.$\uparrow$} & {\small Lay.$\uparrow$} &
  {\small \#Obj$\uparrow$} & {\small \#OB$\downarrow$} & {\small \#CN$\downarrow$} & {\small \#Rch$\uparrow$} & {\small Real.$\uparrow$} & {\small Div.$\uparrow$} & {\small Lay.$\uparrow$} \\
\midrule
LayoutGPT & 8.3 (0.0) & 0.1 & 2.6 & 95.3 & 3.9 & 33.3 & 7.8 &
           14.7 (0.0) & 0.0 & 0.5 & 98.6 & 3.9 & 3.9 & 13.7 \\
Holodeck  & 17.5 (7.1) & 0.0 & 0.0 & 88.7 & 41.2 & 9.8 & 39.2 &
           \textbf{64.4 (25.2)} & 0.0 & 0.0 & 80.0 & 0.0 & \textbf{51.0} & 5.9 \\
\textbf{Ours} & \textbf{22.6 (9.2)} & \textbf{0.0} & \textbf{0.0} & \textbf{100.0} & \textbf{54.9} & \textbf{56.9} & \textbf{52.9} &
                57.3 (19.8) & \textbf{0.0} & \textbf{0.0} & \textbf{100.0} & \textbf{96.1} & 45.1 & \textbf{80.4} \\
\midrule
\multirow{2}{*}{Method} &
\multicolumn{7}{c}{Restaurant (polygon)} &
\multicolumn{7}{c}{Library (polygon)} \\
\cmidrule(lr){2-8}\cmidrule(l){9-15}
& {\small \#Obj$\uparrow$} & {\small \#OB$\downarrow$} & {\small \#CN$\downarrow$} & {\small \#Rch$\uparrow$} & {\small Real.$\uparrow$} & {\small Div.$\uparrow$} & {\small Lay.$\uparrow$} &
  {\small \#Obj$\uparrow$} & {\small \#OB$\downarrow$} & {\small \#CN$\downarrow$} & {\small \#Rch$\uparrow$} & {\small Real.$\uparrow$} & {\small Div.$\uparrow$} & {\small Lay.$\uparrow$} \\
\midrule
LayoutGPT & 7.7 (0.0) & 1.0 & 2.5 & 89.6 & 3.9 & 5.9 & 19.6 &
           13.3 (0.0) & 1.0 & 3.7 & 97.2 & 11.8 & 9.8 & 9.8 \\
Holodeck  & 74.4 (21.8) & 0.0 & 0.0 & 65.2 & 15.7 & 37.3 & 11.8 &
           73.6 (36.6) & 0.0 & 0.0 & 88.3 & 5.9 & 5.9 & 9.8 \\
\textbf{Ours} & \textbf{78.1 (25.6)} & \textbf{0.0} & \textbf{0.0} & \textbf{100.0} & \textbf{80.4} & \textbf{56.9} & \textbf{68.6} &
                \textbf{88.6 (34.3)} & \textbf{0.0} & \textbf{0.0} & \textbf{100.0} & \textbf{82.4} & \textbf{84.3} & \textbf{80.4} \\
\bottomrule
\end{tabular*}
\end{table*}

The experimental results are shown in Table~\ref{tab:single_room_object}. Our method achieves lower collision rates and higher reachability while maintaining a high number of placed objects across different room types, achieving 100\% reachability in all scenes. The user study further shows that our method performs better in overall layout quality and visual realism.

\begin{figure}[t]
\centering
\includegraphics[width=\linewidth]{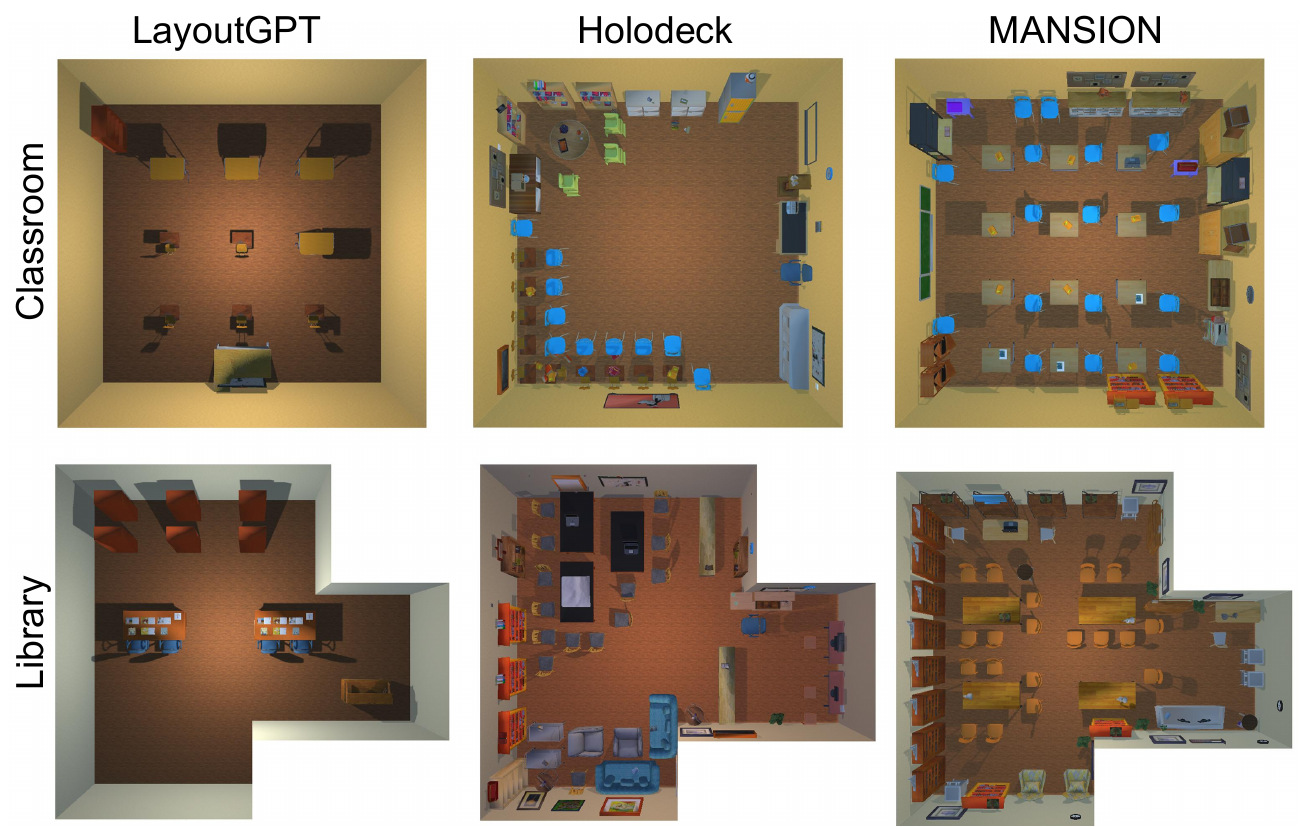}
\caption{object placement qualitative comparison.}
\label{fig:objectexp_qual}
\end{figure}

This advantage is particularly evident in non-residential environments such as classrooms, libraries, and offices. Compared with residential scenes, these environments typically involve larger spaces, higher object density, and stronger demands for regular arrangements, as illustrated in Fig.~\ref{fig:objectexp_qual}. One takeaway from the user study is that MANSION slightly underperformed on Classroom in terms of object count and diversity. We hypothesize that this is mainly because classroom layouts contain a large number of identical desks and chairs arranged in regular, repetitive rectangular formations, which improves structural regularity and reachability but reduces the perceived diversity of the layout.

\subsection{Embodied algorithms in MANSION}
\label{sec:algo_man}
To further explore the downstream applications of MANSION, we validate its effectiveness by cross-implementing BUMBLE \citep{shah2025bumble}, COME-robot \citep{zhi2025closed}, and a variant of BUMBLE with text augmentation. We postpone the introduction of these algorithms and skill library adaptations until Supp.~\ref{sec:embodied_algo}. We design long-horizon, cross-floor tasks to evaluate system performance across three settings: 1) single-floor apartment environments, 2) two-floor office environments connected by stairs or an elevator, and 3) a four-story building with an elevator. Following the object-retrieval setup of \citet{shah2025bumble}, each task requires the agent to navigate the environment to locate a target object. To increase complexity, we add a second delivery phase in which the agent must transport the retrieved object to a specified destination, demanding longer-horizon reasoning. To better understand the sources of failure, we also report a progress score that decomposes overall task completion into two components: successful target object retrieval and successful navigation to the final goal location. Representative failure cases are described in Supp.~\ref{sec:fail}.

\begin{table}[t]
    \centering
    \caption{Task success rates from 10 trials. Progress score is reported in the brackets in the format (Object retrieval success, Navigation success) }
    \label{tab:bumble}
    \begin{tabular}{llll}
        \toprule
    \textbf{Method}                     & \textbf{Single fl.} & \textbf{Double fl.} & \textbf{Four fl.} \\
        \midrule
        COME ~\citep{zhi2025closed}               & 30 (50, 30) & 20 (50, 20)& 0 (0, 0) \\
        BUMBLE ~\citep{shah2025bumble}                & 40 (50, 40) & 20 (30, 40) & 0 (0, 0) \\
        \makecell[l]{\textbf{BUMBLE} \\ \quad w/ object type } & 60 (90, 70) & 60 (80, 60)& 0 (0, 0) \\            
        
        \bottomrule
    \end{tabular}
\end{table}

Table \ref{tab:bumble} summarizes task success rates across 10 trials for single-, double-, and four-floor settings, with progress scores indicating object-retrieval and navigation success. While COME and standard BUMBLE achieve limited performance, adding object-type information to BUMBLE substantially improves success on one- and two-floor tasks. All methods fail in the four-floor setting, reflecting the difficulty of long-horizon, multi-floor tasks. This is probably due to the high-level planner being overwhelmed with information. Across all experiments, we can identify that vision and memory play complementary roles in long-horizon tasks. Enhanced vision modules improve object identification and retrieval success, while memory supports long-horizon navigation by tracking visited locations and avoiding redundant exploration. Together, they are essential for effective multi-floor mobile manipulation. The results also highlight the need to develop new algorithms tackling long-horizon robotics tasks.


\section{Conclusion}
We presented MANSION, a language-driven framework that generates multi-floor, building-scale 3D environments from natural-language descriptions via semantically grounded, vertically aligned floorplans. Built on this framework, we release MansionWorld, a reusable dataset and ecosystem that extends AI2-THOR with cross-floor assets, skill APIs, and task-semantic editing to support long-horizon, multi-floor embodied tasks on shared building layouts. Experiments indicate that our generated floorplans are structurally and functionally reasonable, while existing embodied algorithms still exhibit substantial headroom on MANSION, highlighting the simulation’s value as a testbed for future research.

\newpage
\section*{Acknowledgement}
\label{sec:acknowledge}
We sincerely thank the colleagues and friends (in alphabetical order) from AgiBot, Cornell University, Fudan University, Huazhong Agricultural University, Hunan University, L2S-CentraleSupélec, McGill University, NTU Singapore, Ocean University of China, Peking University, Purdue University, Queen's University, Shanghai Jiao Tong University, Shanghai University of Finance and Economics, SINTEF Ocean, Texas A\&M University, Tsinghua University, University of Delaware, University of Massachusetts Lowell, University of Minnesota Twin Cities, University of Pennsylvania, and University of Science and Technology of China for their participation and support in the user study survey. We are grateful to our colleagues at AgiBot for their meaningful discussions during the preparation of this manuscript.

{
    \small
    \bibliographystyle{ieeenat_fullname}
    \bibliography{main}
}

\clearpage
\section*{Supplemental Materials}
\appendix


\section{Additional Qualitative Results}
\label{sec:appendix-qualitative}

\subsection{Qualitative Floorplan Comparison}
\label{sec:comparison-chd}
\begin{figure}[h]
  \centering
  \includegraphics[width=\linewidth]{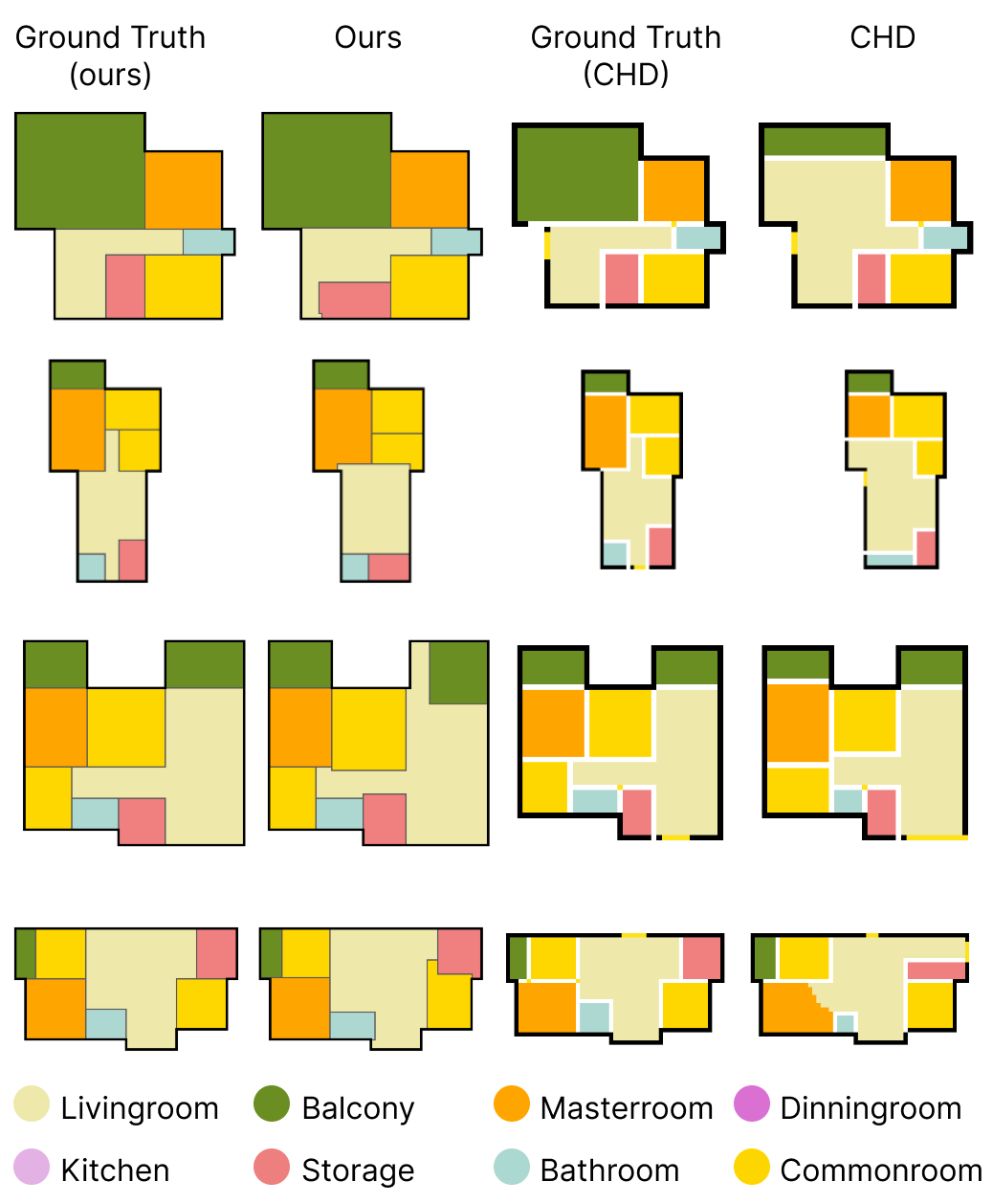}
  \caption{\textbf{Qualitative floorplan comparison with CHD.}}
  \label{fig:comparison-chd}
\end{figure}

We provide a qualitative comparison with ChatHouseDiffusion (CHD)~\cite{qin2024chathousediffusion} under the same raster-space protocol used for the quantitative IoU evaluation in Sec.~4.1, as shown in Fig.~\ref{fig:comparison-chd}. This is important because CHD formulates floor-plan generation as an image-based diffusion process, whereas our method is a training-free geometric solver that operates directly on polygons.

\textbf{Shared raster-space protocol.}
To make the comparison as fair as possible, both methods are evaluated on a common $64 \times 64$ grid aligned to the same floor-plan bounding box. Since CHD produces floor-plan outputs in image form, our method is discretized into the same raster space after uniformly scaling the polygonal layout and quantizing the coordinates, following the evaluation protocol used in Sec.~\ref{sec:floor_gen}.

\textbf{Visual style mismatch.}
The contour mismatch mainly arises from rendering conventions rather than layout structure. CHD visualizations typically include thick exterior contours and explicit white wall gaps between adjacent rooms, whereas our rasterization directly assigns each interior pixel to a room label. As a result, the rendered appearances may differ even when the underlying room arrangements are highly similar.

\textbf{Why the rasterization comparison is fair.}
This rasterization introduces a small representational gap for our method, since a vectorized layout must be converted into raster form before comparison. Minor discrepancies may therefore appear near thin boundaries, corners, or narrow room connections. However, this effect is limited and systematic: both predictions and ground-truth annotations are compared after the same scaling and rasterization procedure, so the quantization error is limited and does not materially affect the IoU comparison.

\subsection{Qualitative comparison with Holodeck}

We further provide a qualitative comparison with existing methods that generate both room layouts and instantiated 3D scenes from high-level semantic descriptions. Representative systems in this category include AnyHome~\cite{fu2024anyhome} and Holodeck~\cite{Yang_2024_Holodeck}. We focus on Holodeck, since AnyHome relies on a HousGAN++-style floorplan backend mainly tailored to residential settings, making it less compatible with our open-vocabulary, non-residential scenario.

\begin{figure*}[t]
  \centering
  \includegraphics[width=\textwidth]{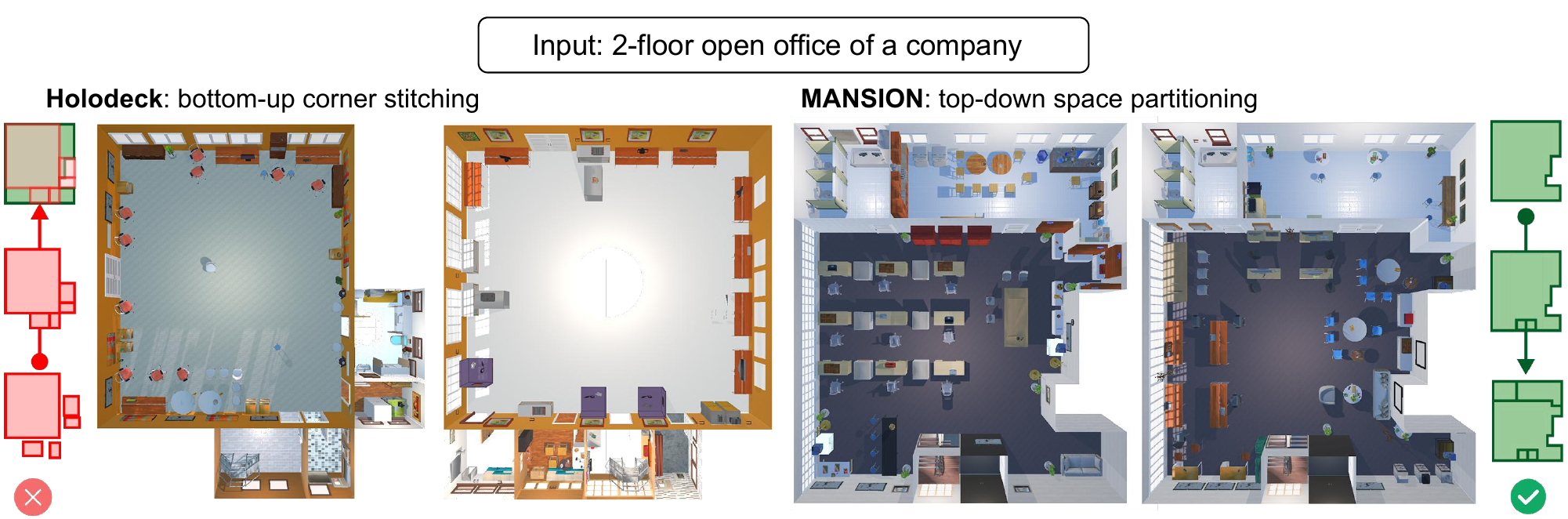}
  \caption{Qualitative comparison between Holodeck and MANSION under high-level semantic building prompts.}
  \label{fig:holodeck_compare}
\end{figure*}

Figure~\ref{fig:holodeck_compare} highlights a key methodological difference. Holodeck follows a bottom-up paradigm that directly predicts room corners and stitches them into a floor layout. This design is effective for lightweight single-floor scene synthesis, but it does not explicitly model building contours, topology-preserving floor partitioning, or cross-floor structural consistency. In contrast, MANSION adopts a top-down formulation, where each floor is generated as a constrained partition under contour, topology, and vertical-core constraints. This makes our method better suited for multi-floor buildings and large-scale non-residential spaces.

As shown in Fig.~\ref{fig:holodeck_compare}, this top-down design provides five practical advantages: \textbf{contour control}, \textbf{topology control}, \textbf{vertical alignment}, \textbf{realistic placement}, and \textbf{style consistency}. The first three stem from our constrained floorplan formulation, while the latter two come from our scene-instantiation design for large, non-residential spaces and building-level room-card style propagation.

The goal of this comparison is to contrast generation paradigms rather than claim strict metric superiority under a shared benchmark. Since Holodeck only produces corner-based layouts and does not explicitly target contour or topology controllability, the most faithful comparison at the layout-generation level is qualitative.

\subsection{Structural Flexibility and Physical Fidelity}

\begin{figure*}[t]
  \centering
  \includegraphics[width=\textwidth]{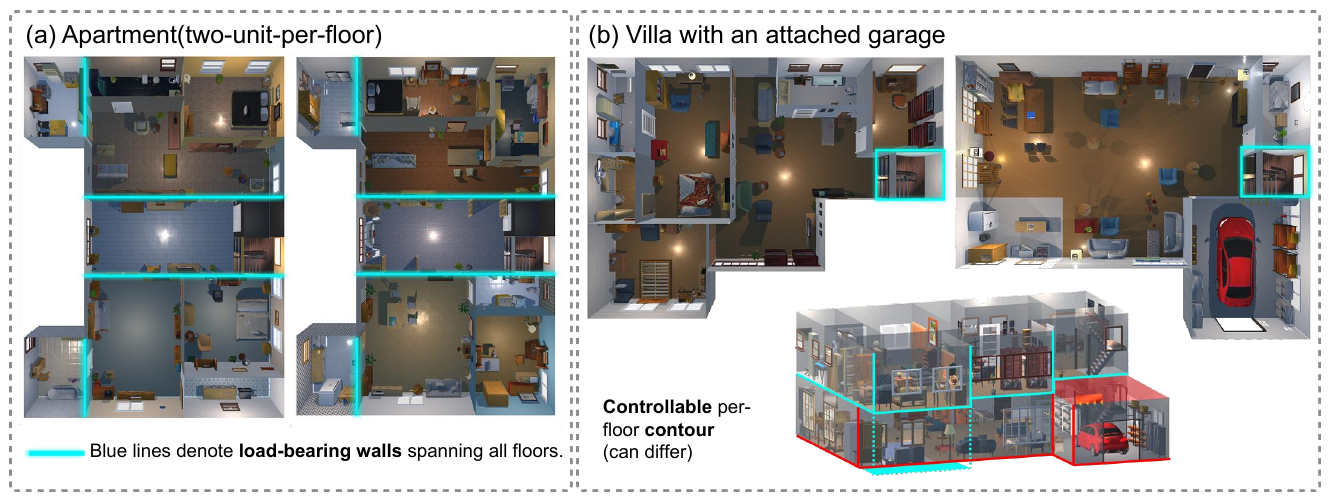}
  \caption{Structural realism in MANSION.
  (a) Cross-floor load-bearing walls (blue) are preserved to maintain vertical consistency.
  (b) A villa with a protruding room and garage shows controllable per-floor outer contours.}
  \label{fig:diverse_buildings}
\end{figure*}

Fig.~\ref{fig:diverse_buildings} highlights two properties of MANSION that are important for physical realism. First, MANSION is \emph{contour-controllable} rather than restricted to identical cross-floor outlines. The consistent contours used in some main-paper experiments are an evaluation simplification, not an algorithmic requirement: floor-wise footprints may differ across floors when specified by the building program.

Second, our notion of vertical consistency goes beyond stairs and elevators. During recursive partitioning, load-bearing walls and other fixed vertical structures are preserved as geometric constraints for subsequent floor splits, enabling cross-floor structural coherence in buildings with more realistic and complex organization. As a result, MANSION  supports better apartment-, office-, and villa-style structures that are closer to real-world buildings, rather than simple floor-by-floor stacking.


\section{MansionWorld Dataset Details}
\subsection{Physical Scale and Functional Composition}

To build a benchmark environment that both supports high-performance simulation and enables a comprehensive evaluation of embodied AI across tasks of varying complexity, \textsc{MansionWorld} is carefully designed along two dimensions: physical scale and functional scene composition.

Considering the physics load of AI2-THOR when handling dense rigid-body interactions, we cap the effective area of each floor at about $500\,\text{m}^2$ to maintain stable frame rates in complex interaction scenes. Leveraging the dynamic floor loading mechanism of the \textsc{MANSION} framework, we adopt a \emph{single-floor constrained, vertically open} spatial strategy: while the area of each individual floor is kept within a controlled range for simulation efficiency, the total number of floors in a building can be extended up to ten.

For functional composition, instead of uniformly sampling scene types, we follow the major application domains of current real-world robots and construct a three-way mixture of \emph{residential} (50\%), \emph{office} (30\%), and \emph{public} (20\%) buildings. This mixture is intended to cover home service robots, intra-building delivery and inspection robots, as well as robots operating in public spaces such as shopping malls, hospitals, and campuses. While residential scenes form roughly half of the corpus in order to support an easy-to-hard curriculum grounded in everyday household tasks, a key novelty of \textsc{MansionWorld} compared to prior home-centric benchmarks lies in its substantial share of non-residential office and public buildings at the building scale. These non-residential environments are where most of our long-horizon, building-scale evaluations are conducted, and they underpin the ``non-residential'' emphasis in the main paper.

On top of this, we deliberately impose a \emph{difficulty curriculum} where simple scenes are more frequent while complex scenes form a long tail. Residential buildings serve as the basic testbed: a large number of \emph{Studio \& Small Flat} units, although compact in size (typically $60\text{--}90\,\text{m}^2$), are populated with high object density and intentionally irregular layouts, in order to stress-test agents' fine-grained manipulation, short-range navigation, and robustness to clutter (e.g., avoiding toys in a messy living room to find a TV remote). In contrast, multi-floor \emph{Family Apartment} and \emph{Duplex \& Townhouse} units introduce vertical connections via internal staircases and elevators, enabling cross-floor tasks in domestic environments. Agents must explicitly model the abstract notion of ``floor'' to accomplish compound household tasks that depend on spatial memory and state tracking, such as \emph{``collect dirty clothes from the bedroom on the second-floor and bring them to the laundry room on the first-floor.''}

Office and public buildings further emphasize semantic reasoning and socially aware navigation. The office subset often exploits large, nearly $500\,\text{m}^2$ floor plates with long corridors and repetitive workstation patterns, posing challenges for robust self-localization in highly similar local structures and for long-range intra-building delivery (e.g., distributing documents or parcels across an eight-floor building). The public subset (e.g., shopping malls, hospitals, schools) highlights explicit functional zoning and semantic priors: agents cannot rely on geometry alone, but must leverage commonsense knowledge such as ``pharmacies are not located in cafeterias'' or ``fresh produce sections tend to be adjacent to cold-chain facilities'' to build high-quality semantic maps and perform efficient target search. This addresses a gap in existing datasets, which mostly focus on homes and single-floor dwellings. Overall, \textsc{MansionWorld} contains a larger number of low-rise, small-to-medium scale scenes that are convenient for day-to-day algorithm development and rapid evaluation, while still reserving a non-trivial proportion of high-rise, large-scale office and public buildings to stress-test the generality and upper-limit performance of embodied systems.

\begin{figure}[ht]
    \centering
    \includegraphics[width=0.85\linewidth]{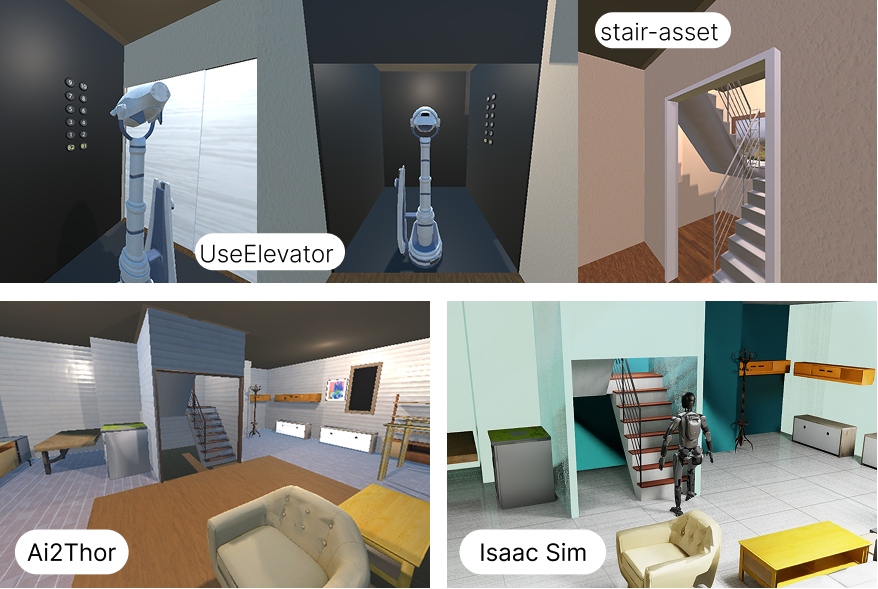}
    \caption{Additional details of the MansionWorld ecosystem.}
    \label{fig:crossfloor_transfer}
\end{figure}

\subsection{Qualitative Examples of MansionWorld Scenes}

\begin{figure*}[t]
  \centering
  \includegraphics[width=\textwidth]{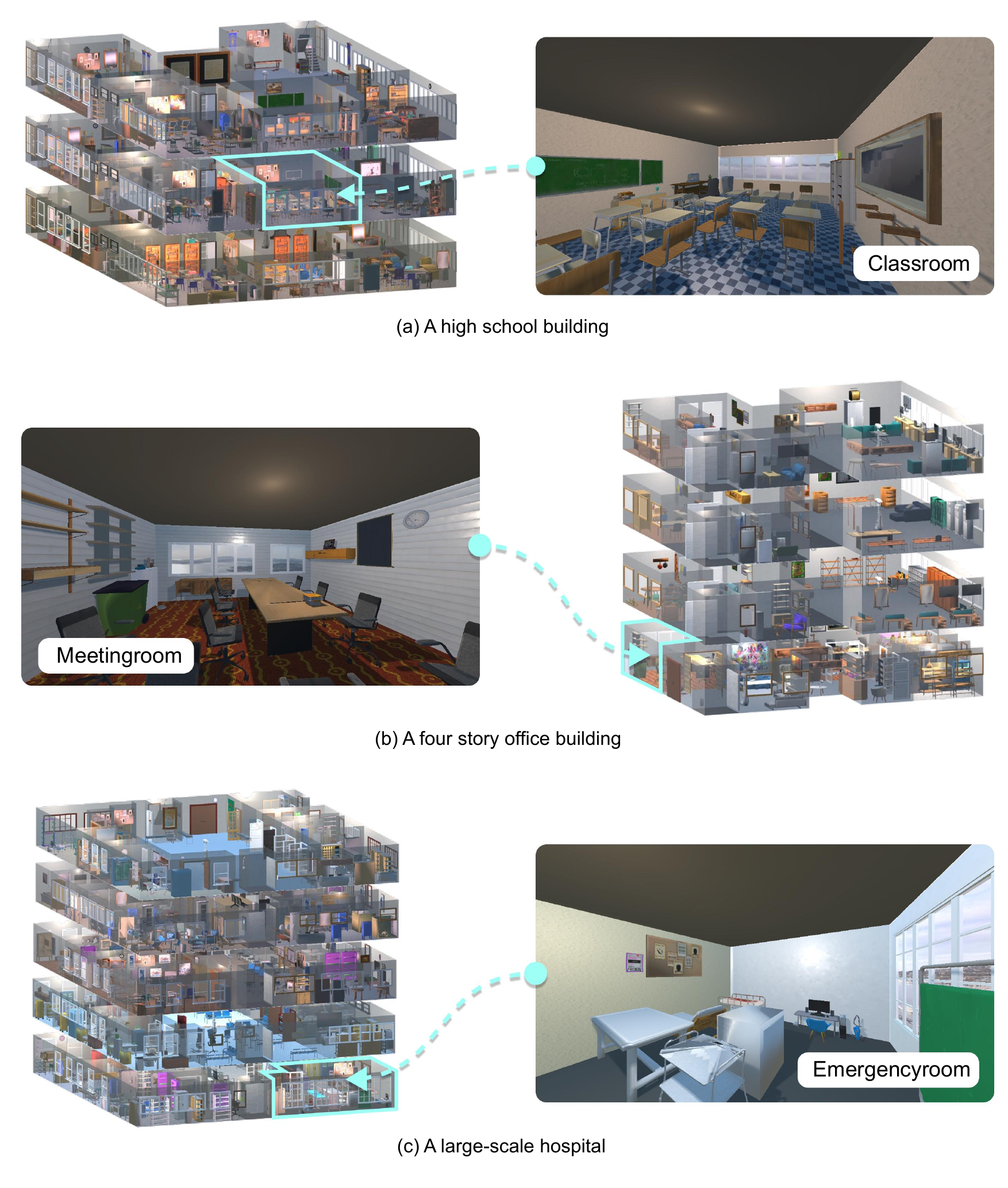}%
  \caption{
  Qualitative examples of non-residential buildings in \textsc{MansionWorld}.}

  \label{fig:mansionworld_examples_nonres}
\end{figure*}

\begin{figure*}[t]
  \centering
  \includegraphics[width=\textwidth]{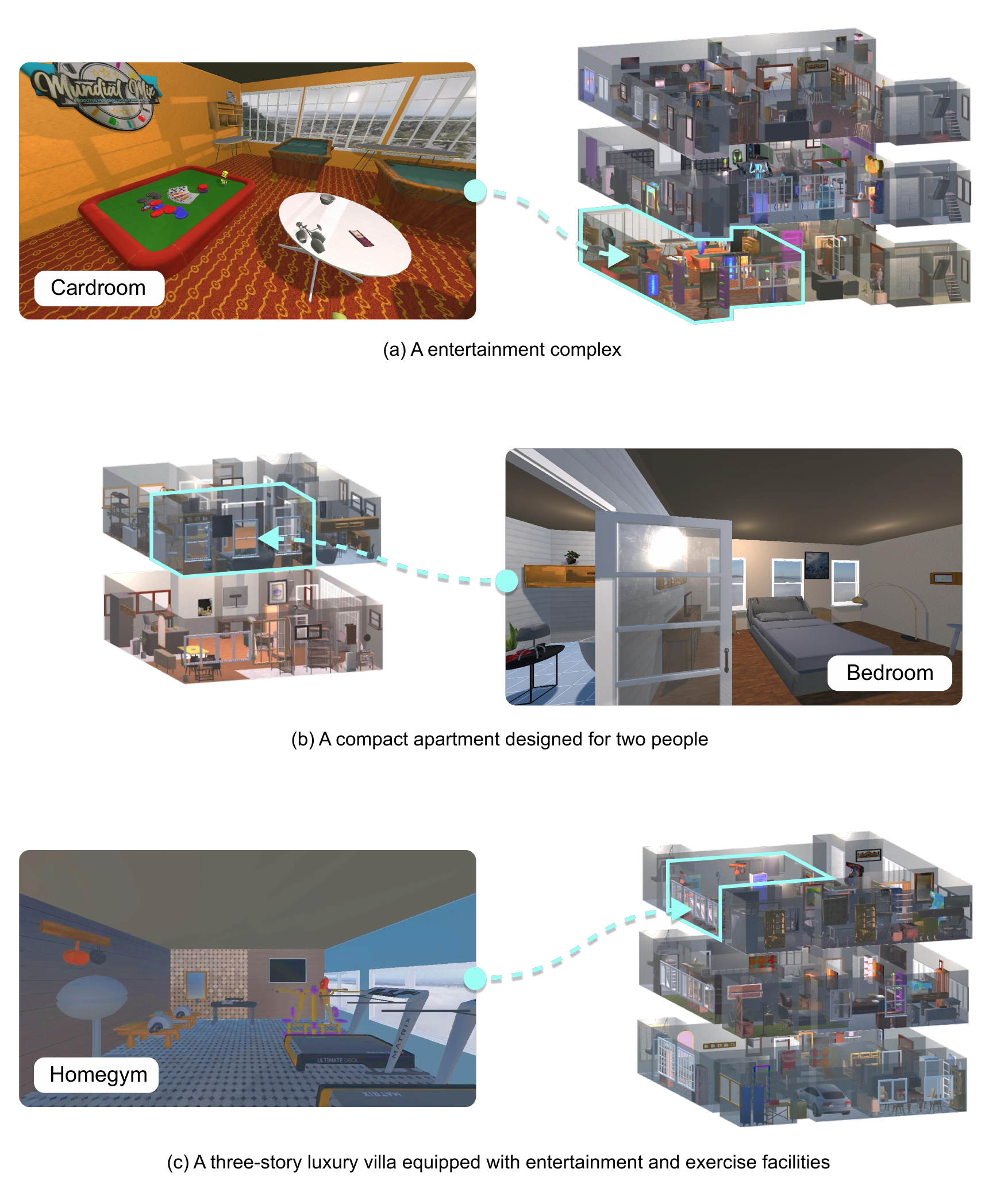}%
  \caption{
  Qualitative examples of entertainment and residential buildings in \textsc{MansionWorld}.}
  \label{fig:mansionworld_examples_res}
\end{figure*}

To complement the statistics in Fig.~\ref{fig:dataset}, we further visualize several representative buildings from \textsc{MansionWorld} and the egocentric observations perceived by an embodied agent operating inside these buildings. Each example pairs a 3D view of a multi-floor building with a first-person view from a highlighted room, see Fig.~\ref{fig:mansionworld_examples_nonres} and Fig.~\ref{fig:mansionworld_examples_res}.

\subsection{Cross-Floor Mobility and Simulator Transfer}
\label{sec:appendix_crossfloor_transfer}

Fig.~\ref{fig:crossfloor_transfer} shows several of our extended cross-floor assets and skills, which support floor-to-floor interaction in MansionWorld, as well as an example of transferring an AI2-THOR scene to NVIDIA Isaac Sim.

\section{Multi-floor Generation Pipeline Details}
\label{sec:appendix-pipeline}

\subsection{Input specification and global orchestration}

The user (or a higher-level generator) provides a
natural-language building description $D$ together with a (possibly
partial) set of numerical constraints. In practice, most of these
constraints can be inferred by a large language model from $D$, and
only the geometric footprint is synthesized by the planner. For
clarity, we write them explicitly as
\begin{itemize}
  \item \textbf{Target floor count} $F_{\text{target}}$ (optional):
        a desired number of floors. When not explicitly specified, it
        is inferred from $D$ (e.g., ``two-storey townhouse'' or ``high-rise office'').
  \item \textbf{Target floor area} $A_{\text{target}}$ (optional):
        a desired gross floor area (per building or per floor). If not
        given, it is similarly inferred from $D$ under simulator
        constraints (e.g., a per-floor area cap for stable physics).
  \item \textbf{Footprint constraint} $P_{\text{env}}$ (optional):
        an outer polygon of the building envelope. In
        our main experiments, this footprint is \emph{not} provided by
        the user; instead, the planner samples a feasible outline
        consistent with $(F_{\text{target}}, A_{\text{target}})$ and
        engine limits, and we denote the resulting footprint as
        $P_{\text{env}}$.
\end{itemize}
Any of the scalar constraints $F_{\text{target}}$ and
$A_{\text{target}}$ may be omitted in the input; in that case, the
planner first invokes an LLM to parse $D$ and derive reasonable default
values. The footprint $P_{\text{env}}$ is typically synthesized (or, in
dataset-driven settings, supplied by the benchmark) and is never
manually drawn by the user in our pipeline. This makes the system
applicable both when the user prescribes an approximate scale (``three
small floors'') and when global dimensions are left entirely to the
generator.

Given $(D, F_{\text{target}}, A_{\text{target}}, P_{\text{env}})$, the
multi-floor controller first synthesizes a global building program
$B_{\text{plan}}$ as above. It then proceeds floor by floor. For each
floor index $i \in \{1,\dots,F\}$, it generates a symbolic room topology
$G_i = (R_i, E_i)$ consistent with the cross-floor skeleton, and calls 
the single-floor solver in Algorithm~\ref{alg:floor-solver} to turn $G_i$
into a geometric layout $L_i$. Internally, this solver constructs a
cut schedule $\mathcal{R}_i$ and applies the topology-aware cutting node
in Algorithm~\ref{alg:cut-node} round by round. Once the 2D floorplan
$L_i$ is fixed, a deterministic instantiation pipeline applies floor
surfaces, walls and openings, then iterates over rooms to populate large
and small objects, and finally adds lighting, skybox and agent spawns to
obtain an executable 3D scene. The whole process is summarized below.

\begin{algorithm}[t]
  \caption{Global multi-floor generation pipeline}
  \label{alg:multifloor-pipeline}
  \begin{algorithmic}[1]
    \Require Description $D$; optional $F_{\text{target}}$, $A_{\text{target}}$, $P_{\text{env}}$
    \Ensure Per-floor scenes $\{S_i\}_{i=1}^{F}$
   \State $(\hat{F}, \hat{A}) \gets 
    \textsc{ResolveNumericConstraints}(D, $ \\
\hspace*{15em} $\backslash F_{\text{target}}, A_{\text{target}})$
    \State $B_{\text{plan}} \gets \textsc{PlanBuildingProgram}(D, \hat{F}, \hat{A}, P_{\text{env}})$
    \State $F \gets \textsc{NumFloors}(B_{\text{plan}})$
    \State $S \gets \emptyset$
    \For{$i = 1$ to $F$} \label{line:floor-loop}
      \State $G_i \gets \textsc{GenerateFloorTopology}(B_{\text{plan}}, i)$
      \State $L_i \gets \textsc{SolveFloorLayout}(G_i, B_{\text{plan}}, i)$
      \State $X_i \gets \textsc{ApplyFloorStructure}(L_i)$
      \State $Y_i \gets \textsc{ApplyWallsAndOpenings}(X_i)$
      \For{each room $r \in \textsc{Rooms}(L_i)$} \label{line:room-loop}
        \State $Y_i \gets \textsc{PlaceLargeObjects}(Y_i, r)$
        \State $Y_i \gets \textsc{PlaceSmallObjects}(Y_i, r)$
      \EndFor
      \State $Y_i \gets \textsc{AddLighting}(Y_i)$
      \State $Y_i \gets \textsc{AddSkybox}(Y_i)$
      \State $S_i \gets \textsc{PlaceAgentSpawn}(Y_i, B_{\text{plan}}, i)$
      \State $S \gets S \cup \{S_i\}$
    \EndFor
    \State \Return $S$
  \end{algorithmic}
\end{algorithm}

\begin{figure}[t]
  \centering
  \includegraphics[width=\linewidth]{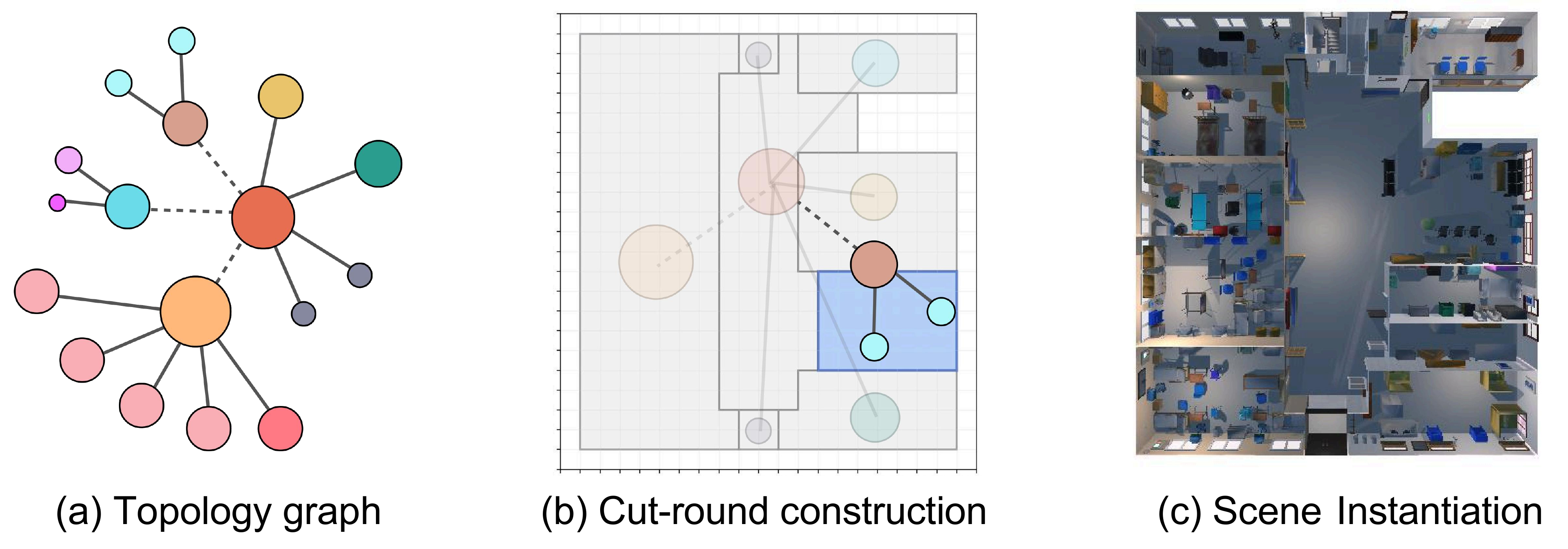}
  \caption{
  Illustration of the single-floor, topology-driven pipeline.
  (a) Input room topology graph.
  (b) Cut-round construction and hierarchical splitting over the free region.
  (c) Final 3D scene instantiation in AI2-THOR after applying structure, objects, and lighting.
  }
  \label{fig:topology_cut_instantiation}
\end{figure}

\subsection{Single-floor topology-driven floorplan solver}
\label{sec:appendix-floor-solver}
Fig.~\ref{fig:topology_cut_instantiation} illustrates the single-floor
solver: given a symbolic room topology graph, we construct cut rounds
over the free region and finally instantiate the resulting layout as an
executable 3D scene.
We reuse the notation from Sec~\ref{sec:framework}: for floor $f$ we write
$\Omega_f$ for the free region after removing vertical cores and
$G_f = (R_f, E_f)$ for the room graph with target areas
$\{a_r\}_{r \in R_f}$. The solver is a constructive procedure
that approximately optimizes the layout objective in Sec.~3.1
under the hard topological constraint $\mathrm{Topo}(L,G_f)$.

The algorithm proceeds by hierarchical splitting. A hub node
$\mathit{main} \in R_f$ is chosen as the root; a cut-planning
routine constructs a sequence of rounds
$\mathcal{R}_f = \{(p_t, C_t)\}_{t=1}^T$ with parents
$p_t \in R_f$ and non-empty child sets $C_t \subseteq R_f$; and
a generic cutting node successively refines the layout in each
round.

\begin{algorithm}[t]
  \caption{Single-floor topology-driven floorplan solver}
  \label{alg:floor-solver}
  \begin{algorithmic}[1]
    \Require Floor index $f$; free region $\Omega_f$; room graph $G_f = (R_f, E_f)$;
             target areas $\{a_r\}_{r \in R_f}$; vertical cores $V$
    \Ensure Floorplan layout $L_f$ partitioning $\Omega_f$
    \State $\mathit{main} \gets \textsc{SelectHubNode}(G_f)$
    \State $\mathcal{R}_f \gets \textsc{BuildCutRounds}(G_f, \mathit{main}, V)$
    \State $L_f \gets \textsc{InitLayout}(\Omega_f, V, \mathit{main})$
    \For{\textbf{each} $(p_t, C_t) \in \mathcal{R}_f$}
      \State $L_f \gets \textsc{CutNode}\big(L_f, p_t, C_t, G_f, \{a_r\}\big)$
    \EndFor
    \State \Return $L_f$
  \end{algorithmic}
\end{algorithm}

\paragraph{Cut-round construction.} $\textsc{BuildCutRounds}$ performs a breadth-first
traversal on $G_f$ rooted at $\mathit{main}$, assigns a depth to
each node, and groups non-vertical rooms by depth and parent.
Vertical-core nodes are excluded from the parent set. For each
non-vertical parent $p$ and its non-empty child cluster
$C \subseteq R_f$, it emits a round $(p,C)$. The resulting
$\mathcal{R}_f$ induces an order that respects the graph
structure (no child is instantiated before its parent) and
expands from the hub to the periphery.

\paragraph{Topology-aware cutting node and adaptive growth.}
For a fixed round $(p_t, C_t)$ and current layout $L_f$, the
cutting node first extracts the parent region
$\Omega_f(p_t) \subseteq \Omega_f$ and renders a top-down
preview in which the polygon of $p_t$ is highlighted against the
rest of the floorplan. This image, together with $L_f$, $G_f$,
$p_t$, $C_t$ and $\{a_r\}$, is given to an MLLM that outputs a
seed plan
$\sigma_t = \{(r, c_r, \alpha_r) \mid r \in C_t\}$, where $c_r$
is a continuous seed (approximate centroid) in $\Omega_f(p_t)$
and $\alpha_r$ is a target area fraction consistent with $a_r$
and $|\Omega_f(p_t)|$.

Conditioned on $\sigma_t$, the node runs an adaptive sampling
procedure with $N_{\text{retry}} = 10$ retries and batch size
$B = 100$ local candidates per retry. For each child
$r \in C_t$ it computes an initial radius
$R_r^{(0)} = r_{\text{base}} + k \cdot a_r / |\Omega_f(p_t)|$
(with fixed $r_{\text{base}} = 2$ in grid units and scaling
factor $k$) and at retry $j$ uses a scaled radius
$R_r^{(j)} = \gamma_j R_r^{(0)}$ for a monotonically increasing
sequence $(\gamma_j)_j$. Intuitively, $R_r^{(j)}$ is the adaptive perturbation radius around the seed for room $r$ in retry $j$, controlling how far candidate seeds may
move away from the MLLM-proposed centroid. In retry $j$, it samples $B$ seed
perturbations inside the discs of radius $R_r^{(j)}$ (with a
minimum separation constraint between seeds), grows $B$ local
candidate partitions of $p_t$, filters them by the predicate
$\mathrm{Topo}(\cdot,G_f)$, and scores the survivors with the
score function $\mathrm{Score}(L; \mathbf{w})$ described below. If at
least one candidate survives in retry $j$, the best-scoring one
is accepted and the retry loop terminates. If all
$N_{\text{retry}}$ retries fail, the node falls back to a Monte
Carlo seeding strategy: seeds are sampled uniformly in
$\Omega_f(p_t)$ in decreasing order of target area, subject to
repulsion, and the same growth, topology filtering and scoring
pipeline is applied.

The cutting node is summarized in the following skeleton.

\begin{algorithm}[t]
  \caption{Topology-aware cutting node with MLLM-guided seeds (skeleton)}
  \label{alg:cut-node}
  \begin{algorithmic}[1]
    \Require Layout $L_f$; parent $p_t$; children $C_t$;
             room graph $G_f$; target areas $\{a_r\}$
    \Ensure Updated layout $L_f'$ where $p_t$ is split into $C_t$
    \State $\Omega_f(p_t) \gets \textsc{LocalFootprint}(L_f, p_t)$
    \State $I_t \gets \textsc{RenderHighlightPreview}(L_f, \Omega_f(p_t), p_t)$
    \State $\sigma_t \gets 
    \textsc{PlanSeedsWithMLLM}(I_t, L_f, G_f, $ \\  
    \hspace*{15em} $\backslash p_t, C_t, \{a_r\})$
    \State $\{R_r^{(0)}\} \gets \textsc{ComputeBaseRadii}(\Omega_f(p_t), C_t, \{a_r\})$
    \State $\mathit{best} \gets \textsc{None}$
    \For{$j = 0$ to $N_{\text{retry}}-1$}
      \State $\tilde{\Sigma}_j \gets \textsc{SampleSeedBatch}(\sigma_t, \{R_r^{(0)}\}, j)$
      \State $\mathcal{L}_j \gets \textsc{GrowCandidates}(\Omega_f(p_t), p_t, C_t, \tilde{\Sigma}_j)$
      \State $\mathcal{L}_j \gets \textsc{FilterByTopology}(\mathcal{L}_j, L_f, G_f)$
      \State $\mathit{cand} \gets \textsc{SelectBestByScore}(\mathcal{L}_j)$
      \If{$\mathit{cand} \neq \textsc{None}$}
        \State $\mathit{best} \gets \mathit{cand}$; \textbf{break}
      \EndIf
    \EndFor
    \If{$\mathit{best} = \textsc{None}$}
      \State $\mathit{best} \gets 
    \textsc{FallbackMonteCarlo}(\Omega_f(p_t), p_t, $ \\
    \hspace*{13em} $\backslash C_t, \{a_r\}, L_f, G_f)$
    \EndIf
    \State $L_f' \gets \textsc{MergeLocalPartition}(L_f, p_t, \mathit{best})$
    \State \Return $L_f'$
  \end{algorithmic}
\end{algorithm}

\paragraph{Energy-based scoring and weight selection.}
We now detail the energy-based $\mathrm{Score}(L;\mathbf{w})$
objective introduced in Eq.~\ref{search}.
For each local candidate layout $L$ of the parent region, we
compute a per-room energy and aggregate into a total energy
$E(L;\mathbf{w})$, from which the score is obtained by negation.

For every child room $r \in C_t$ with realized polygon $P_r$,
target area $a_r$, and seed $c_r$, we extract four raw features:
\begin{itemize}
  \item $f_{\text{ratio}}(r)$ (\texttt{ratio}): relative area error
        $\lvert\,\mathrm{area}(P_r) - a_r\rvert / a_r$;
  \item $f_{\text{seed}}(r)$ (\texttt{seed\_dist}): Euclidean
        distance between the centroid of $P_r$ and the input seed
        $c_r$;
  \item $f_{\text{wall}}(r)$ (\texttt{wall\_contact}): absolute
        length of the boundary intersection between $P_r$ and the
        envelope $\partial\Omega_f(p_t)$ of the parent region,
        i.e.\ $\lvert\partial P_r \cap \partial\Omega_f(p_t)\rvert$;
  \item $f_{\text{corner}}(r)$ (\texttt{extra\_corners}):
        $\max(0,\, n_{\text{int}}(r) - 4)$, where $n_{\text{int}}(r)$
        is the number of non-collinear corners of $P_r$ that do
        \emph{not} lie on $\partial\Omega_f(p_t)$.
\end{itemize}

\noindent
Among these, $f_{\text{ratio}}$, $f_{\text{seed}}$, and
$f_{\text{corner}}$ are \emph{penalty} terms (smaller $\Rightarrow$
better), while $f_{\text{wall}}$ is a \emph{reward} term (larger
$\Rightarrow$ better, since more envelope contact yields more
regular rooms).

To balance heterogeneous scales, we apply a \emph{mixed
normalization} strategy.  Only $f_{\text{seed}}$ undergoes
min--max normalization across the room set within a single
candidate:
\[
  z_{\text{seed}}(r) \;=\;
  \operatorname{clamp}_{[0,1]}\!\left(
    \frac{f_{\text{seed}}(r) - f_{\text{seed}}^{\min}}
         {f_{\text{seed}}^{\max} - f_{\text{seed}}^{\min} + \varepsilon}
  \right),
\]
with $f_{\text{seed}}^{\min} = \min_{r} f_{\text{seed}}(r)$ and
$f_{\text{seed}}^{\max} = \max_{r} f_{\text{seed}}(r)$.
For $f_{\text{ratio}}$ and $f_{\text{corner}}$, we found that using raw values directly provides more stable value differences across
candidates with varying room counts and boundary complexities,
because min--max normalization can compress informative
differences when the candidate set is homogeneous.
Similarly, $f_{\text{wall}}$ enters as a raw value clamped to
$[0,1]$.

The per-room energy contribution is
\[
\begin{aligned}
  e(r) \;=\;&\; w_{\text{ratio}}\, f_{\text{ratio}}(r)
  \;+\; w_{\text{seed}}\, z_{\text{seed}}(r) \\
  &\;+\; w_{\text{corner}}\, f_{\text{corner}}(r) \\
  &\;-\; w_{\text{wall}}\,
        \operatorname{clamp}_{[0,1]}\!\bigl(f_{\text{wall}}(r)\bigr),
\end{aligned}
\]
The dominant $w_{\text{ratio}}$ strongly penalizes area
mismatch; $w_{\text{corner}}$ discourages complex room shapes;
$w_{\text{wall}}$ encourages rooms to align with the building
envelope; and $w_{\text{seed}}$ provides a mild bias toward the
MLLM-proposed centroid.
where penalty terms enter with positive signs (higher
$\Rightarrow$ worse) and the wall reward enters with a negative
sign (more contact $\Rightarrow$ lower energy).
The total energy sums over all child rooms and relates to the
search objective (Eq.~\ref{search}) by negation:
\[
\begin{aligned}
  E(L; \mathbf{w}) &= \sum_{r \in C_t} e(r), \\
  \mathrm{Score}(L; \mathbf{w}) &= -E(L; \mathbf{w}).
\end{aligned}
\]

Among all candidates in a round that satisfy
$\mathrm{Topo}(\cdot,G_f)$, the cutting node retains the layout
with the lowest energy (equivalently, the highest
$\mathrm{Score}$).

We set $\mathbf{w}$ by random search on a held-out subset of the
RPLAN dataset: candidate weight vectors are sampled from a
low-dimensional simplex, the solver is run on RPLAN-style
instances, and configurations that yield good area agreement and
regular room shapes are retained; one such $\mathbf{w}$ is fixed
for all experiments.
We intentionally use this hand-crafted, interpretable energy
rather than a learned scoring network, since RPLAN is dominated
by residential layouts and a learned scorer trained on it would
be strongly domain-specific. In contrast, the feature-based
energy can be reweighted to accommodate different building types
without retraining.

\paragraph{Spur removal and hole filling.}
After growth, room polygons may contain thin protrusions (spurs)---single
cells connected to the room body by at most one edge.
A spur cell is identified as any occupied cell whose same-room
4-neighbor count is at most one while having at least one neighbor
belonging to a different room or lying outside the interior.
Spur removal proceeds iteratively: in each pass, all detected spur cells are set to empty; passes repeat until no spurs remain, yielding a spur-free grid.
Subsequently, each remaining empty connected component within the interior is filled by the room sharing the longest boundary with
that component. This fill-then-clean cycle repeats up to 20
iterations to ensure that hole filling does not re-introduce
spur artifacts.

A practical limitation is that when $\Omega_f(p_t)$ is highly
constrained and $G_f$ is complex, the number of candidates satisfying the hard topological constraints can be very small.
In this scenario, the solver is feasibility-driven, where
the influence of $\mathrm{Score}(L; \mathbf{w})$ is reduced.

\section{Task-Semantic Scene Editing Agent}

\subsection{System Architecture}
\label{sec:arch}
\begin{figure*}[t]               
    \centering
    \includegraphics[width=\textwidth]{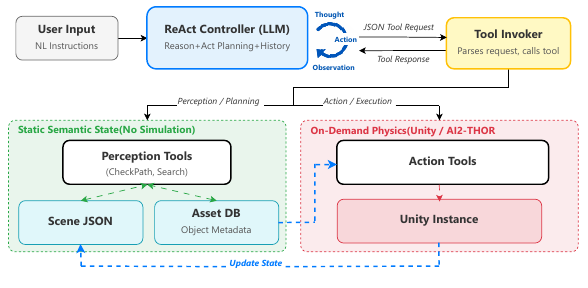} 
    \caption{\textbf{System Architecture of the Task-Semantic Scene Editing Agent.} 
    The system operates via a \textbf{ReAct Controller} (top) that iteratively plans and issues JSON tool requests. 
    A \textbf{Tool Invoker} (middle) serves as an execution bridge, routing perception tasks to the fast 
    \textbf{Static Semantic State} (bottom left) and action tasks to the \textbf{On-Demand Physics Engine} 
    (bottom right). The dashed arrow highlights the \textbf{Hybrid State Management} mechanism, 
    where physical simulation results are synchronized back to the static scene JSON to ensure consistency.}
    \label{fig:agent_arch}
\end{figure*}
We illustrate the detailed architecture of our Task-Semantic Scene Editing Agent in Fig.~\ref{fig:agent_arch}. Operating as a high-level \textbf{neuro-symbolic scheduler}, our system decouples semantic reasoning from physical simulation, achieving both computational efficiency and physical plausibility. The architecture comprises three core subsystems:

\noindent\textbf{The ReAct Controller (Cognitive Layer).}
The agent's core is a Large Language Model (LLM) utilizing a ReAct (Reason+Act) protocol. It processes natural language instructions and conversation history, outputting structured JSON actions to invoke specific tools. This abstraction enables the agent to plan over long horizons by manipulating scene semantics rather than raw pixels or low-level motor commands.

\noindent\textbf{Hybrid State Management (Dual-Backend).}
A key innovation is the separation of static data from dynamic simulation. This dual-backend approach ensures optimal resource utilization:
\begin{itemize}
    \item \textbf{Static Semantic State (JSON + Asset DB):} The primary "source of truth" is a lightweight Holodeck-compatible JSON file. All logical checks (e.g., path connectivity) and geometric planning (e.g., surface area calculation) are executed directly against this JSON structure and an external Asset Metadata Database (Asset DB). This avoids rendering overhead, enabling rapid "mental simulation" and topological reasoning.
    \item \textbf{On-Demand Physics Engine (Unity):} Calling physics simulation is an expensive, on-demand resource. An AI2-THOR controller is temporarily instantiated only for actions requiring physical validation (specifically, \texttt{PlaceInContainer} or \texttt{PlaceOnSurface}). It executes atomic physics-based actions (e.g., \texttt{SpawnAsset}, \texttt{OpenObject}, \texttt{PlaceObjectAtPoint}) to resolve collisions and gravity. The object's final valid pose is then synchronized back to the static JSON, and after that, the simulator instance is stopped.
\end{itemize}

\noindent\textbf{The Tool Invoker.}
Serving as the execution bridge, this component parses the JSON requests from the ReAct controller and redirects function calls to the appropriate backend, from querying the static semantic state for fast perception tasks to triggering the on-demand physics engine for complex interactions, and returns the execution results as observations to the agent.

\subsection{Tool Library Cards}

We present the detailed specifications of the toolset used in the "Check-and-Provision" workflow. For brevity, we categorize the \textbf{Data Source} of each tool into three components:
\begin{itemize}
    \item \textbf{JSON}: Operations on the static scene graph file (fast, geometric).
    \item \textbf{Asset DB}: Queries to the external Objaverse/AI2-THOR metadata library.
    \item \textbf{Unity}: Runtime physics simulation via AI2-THOR (atomic actions).
\end{itemize}

\subsubsection{Perception Tools (Checking Phase)}

\begin{ToolCard}{Tool: CheckPath}
    \ToolKey{Data Source} JSON
    
    \ToolKey{Description} Verifies topological connectivity and room existence to validate navigation feasibility.
    
    \ToolKey{Input} Global Context (No specific arguments)
    
    \ToolKey{Logic} 
    Parses the polygon boundaries of each room and the coordinates of connecting portals (doors/stairs) from the JSON. It constructs a topological path graph to verify if a valid navigable route exists between the task's start and end locations.
\end{ToolCard}

\begin{ToolCard}{Tool: SearchAssets}
    \ToolKey{Data Source} Asset DB
    
    \ToolKey{Description} Retrieves new interactive objects based on natural language queries.
    
    \ToolKey{Input} \texttt{query}, \texttt{top\_k}, \texttt{properties} (optional)
    
    \ToolKey{Logic} 
    Utilizes the retrieval method from \textbf{Holodeck}: it employs CLIP (visual) and SBERT (semantic) embeddings to match the user's query against the asset library. It returns \texttt{assetId}s that match specific physical properties (e.g., \texttt{Pickable}).
\end{ToolCard}

\begin{ToolCard}{Tool: ListObjects}
    \ToolKey{Data Source} JSON + Asset DB
    
    \ToolKey{Description} Lists existing objects in the scene, supporting filtering by location and functional properties.
    
    \ToolKey{Input} \texttt{room} (optional), \texttt{keyword/id} (optional)
    
    \ToolKey{Logic} 
    Iterates through the JSON to find objects. The \texttt{keyword} input supports both fuzzy name matching and exact ID lookup. It cross-references each object's \texttt{assetId} with the Asset DB to retrieve implicit functional properties (e.g., \texttt{Receptacle}, \texttt{CanOpen}) not stored in the scene file.
\end{ToolCard}

\begin{ToolCard}{Tool: CheckSurface}
    \ToolKey{Data Source} JSON
    
    \ToolKey{Description} Inspects what is currently placed \textit{on top} of a specific object.
    
    \ToolKey{Input} \texttt{keyword/id}
    
    \ToolKey{Logic} 
    Identifies the target object (by keyword or exact ID) and retrieves its child nodes from the JSON hierarchy. It geometrically verifies the "on-top" relationship by comparing the child's centroid height ($y$) against the parent object's Axis-Aligned Bounding Box (AABB) top surface.
\end{ToolCard}

\begin{ToolCard}{Tool: SearchContents}
    \ToolKey{Data Source} JSON
    
    \ToolKey{Description} Inspects what is currently stored \textit{inside} a container.
    
    \ToolKey{Input} \texttt{keyword/id}
    
    \ToolKey{Logic} 
    Identifies the target container and retrieves its child nodes. Unlike surface checks, it verifies the "inside" relationship by checking if the child's centroid is strictly contained within the vertical range ($y_{min}, y_{max}$) of the parent's AABB.
\end{ToolCard}

\subsubsection{Action Tools (Provisioning Phase)}

Action tools modify the scene. These tools automatically handle collision avoidance via an internal geometric solver before invoking native AI2-THOR actions for physical consistency.

\begin{ToolCard}{Tool: PlaceInContainer}
    \ToolKey{Data Source} JSON + Asset DB + Unity
    
    \ToolKey{Description} Physically instantiates an asset inside an openable container (e.g., placing a cola inside a fridge).
    
    \ToolKey{Input} \texttt{asset\_id}, \texttt{container\_id}
    
    \ToolKey{Logic} 
    The system first calculates valid non-overlapping $(x, z)$ coordinates using an internal geometric solver. It then triggers an on-demand AI2-THOR instance and executes a sequence of \textbf{native atomic actions}:
    \begin{enumerate}
        \item \textbf{\texttt{OpenObject}}: Fully opens the container to ensure accessibility.
        \item \textbf{\texttt{SpawnAsset}}: Instantiates the asset (loaded from Asset DB) at the planned coordinates.
        \item \textbf{\texttt{PlaceObjectAtPoint}}: Uses the physics engine to verify collisions and settle the object.
    \end{enumerate}
    Finally, the stable pose is captured and written back to the static JSON.
\end{ToolCard}

\begin{ToolCard}{Tool: RemoveObject}
    \ToolKey{Data Source} JSON
    
    \ToolKey{Description} Deletes specific objects from the scene.
    
    \ToolKey{Input} \texttt{object\_id} (optional), \texttt{receptacle\_id} (optional)
    
    \ToolKey{Logic} 
    Directly edits the JSON scene graph. If \texttt{object\_id} is provided, it removes that specific node. If \texttt{receptacle\_id} is provided, it recursively deletes all child nodes associated with that receptacle, effectively clearing the surface.
\end{ToolCard}

\begin{ToolCard}{Tool: PlaceOnSurface}
    \ToolKey{Data Source} JSON + Asset DB + Unity
    
    \ToolKey{Description} Physically instantiates an asset on an open surface (e.g., table, sofa).
    
    \ToolKey{Input} \texttt{asset\_id}, \texttt{receptacle\_id}
    
    \ToolKey{Logic} 
    Similar to container placement, this tool uses an internal 2D bin-packing algorithm to determine the planar position. It then invokes the following atomic actions in AI2-THOR:
    \begin{enumerate}
        \item \textbf{\texttt{SpawnAsset}}: Instantiates the asset from the Asset DB.
        \item \textbf{\texttt{PlaceObjectAtPoint}}: Allows the object to settle naturally under gravity.
    \end{enumerate}
    This process ensures the object rests naturally on uneven surfaces (e.g., sofa cushions) without floating or clipping before updating the JSON.
\end{ToolCard}

\section{Embodied Algorithms in Mansion}
\label{sec:embodied_algo}

Here, we briefly introduce BUMBLE \citep{shah2025bumble}, COME-robot \citep{zhi2025closed}, and a variant of BUMBLE with text augmentation. These algorithms are representative embodied mobile robot systems for long-horizon navigation and manipulation tasks. BUMBLE is a whole-building framework with a VLM-driven reasoning core, and an open-world perception system, integrating parameterized navigation and manipulation skills guided by dual-layer memory for long-horizon planning and recovery ~\citep{shah2025bumble}. COME-robot operates similarly as a closed-loop, open-vocabulary system, exposing perception and execution APIs and using GPT-4V to refine code-level plans from visual feedback iteratively, but without long-term memory ~\citep{zhi2025closed}. We enable the global perception map of COME-robot by providing rich object information in the scene when prompting the VLM planner. 

Within MANSION, we adapt the skill libraries and decision modules from both systems to our multi-floor experimental setting and evaluate their performance in terms of success rate and robustness to complex layouts. We omit intricate real-world robotic manipulation components, such as dexterous grasping, localization, and low-level motor control, and instead focus on evaluating high-level sequential decision-making for task completion. To enable richer scene interaction, we extend the systems with new skills built upon the atomic actions provided by AI2-THOR~\cite{kolve2017ai2}, allowing the agents to operate effectively in multi-floor environments. Furthermore, to enhance exploration capabilities, we introduce a rotation skill that enables the agent to reorient itself and continue searching when the target object is not initially in sight. For consistency and to balance API query time with model performance, we adopt GPT-4.1 as the VLM backbone~\citep{shah2025bumble}. However, a key limitation arises from the VLM’s reduced object identification accuracy in simulated environments. Although the agent can generate coherent action sequences for the retrieval and delivery of the task, it frequently misidentifies the target object, leading to task failure. To mitigate this issue, we introduce a variant of BUMBLE that exposes the object type only during skill selection, providing the agent with just enough semantic guidance to better interpret its surroundings. Importantly, the agent does not receive object-type information when executing the skills. 

\begin{figure*}[ht]
\centering
\begin{tcolorbox}[colframe=black,boxrule=0.8pt,arc=2pt]
\begin{minipage}{\textwidth}
\centering
\includegraphics[width=0.24\textwidth]{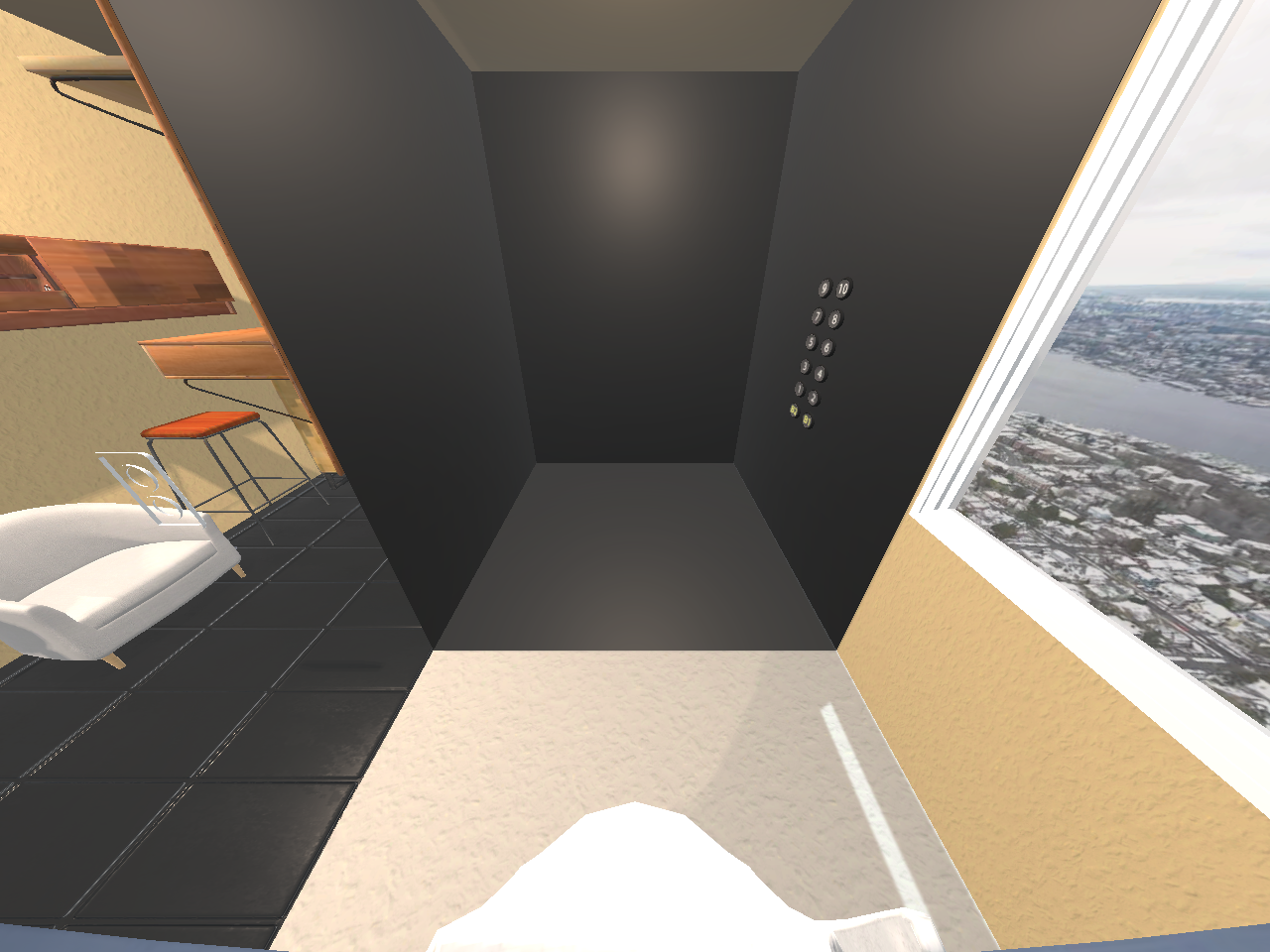}
\includegraphics[width=0.24\textwidth]{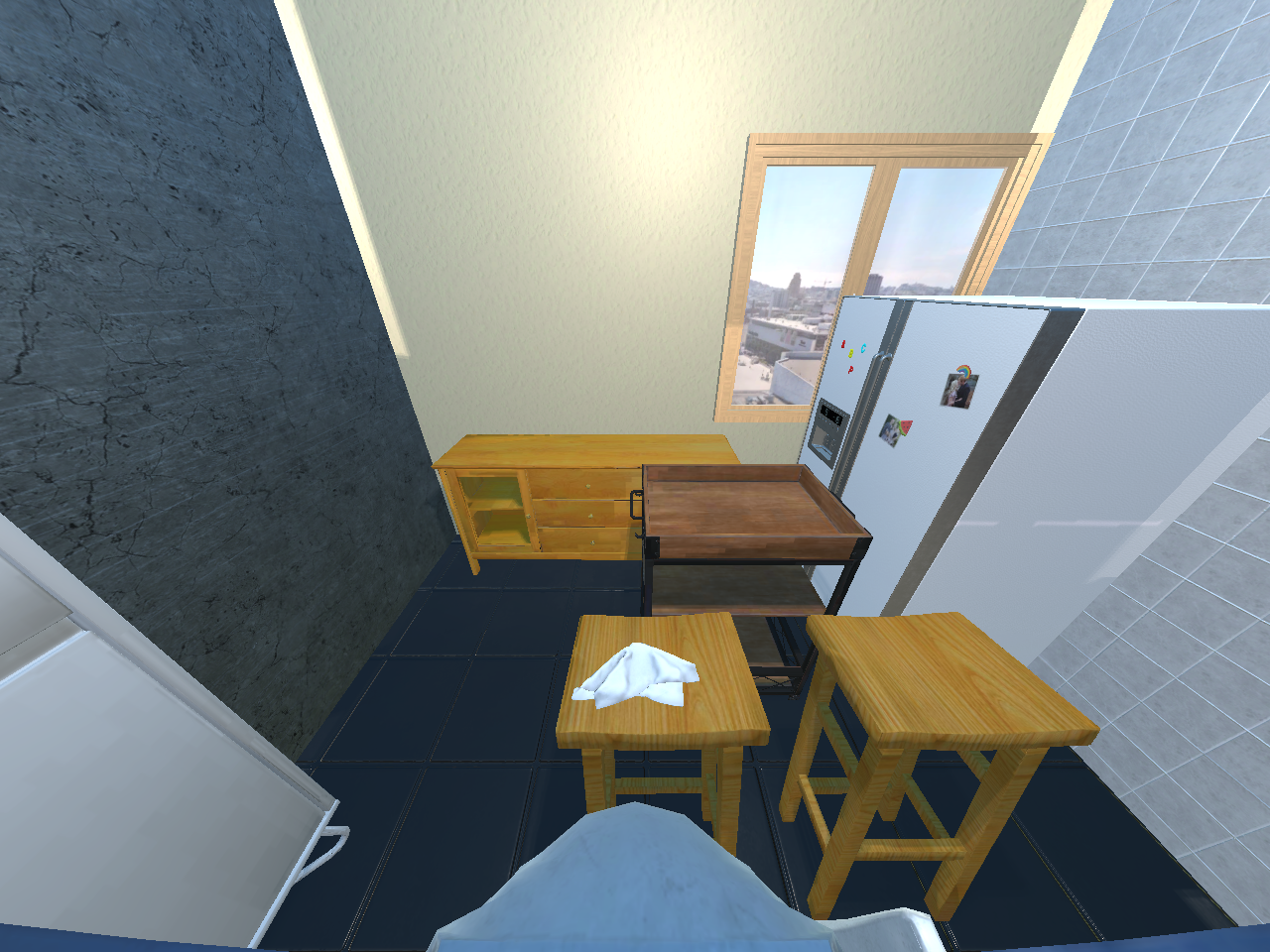}
\includegraphics[width=0.24\textwidth]{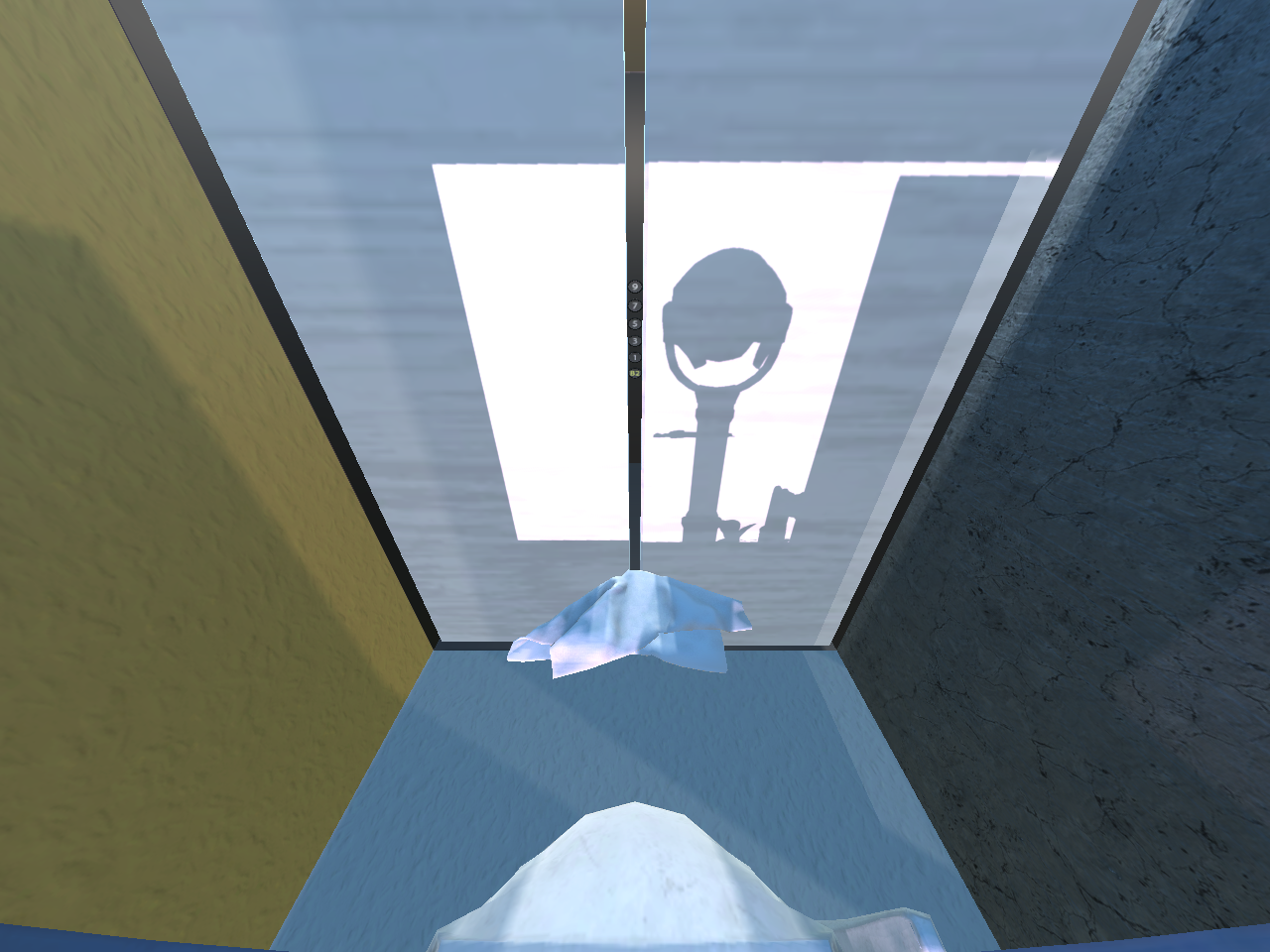}
\includegraphics[width=0.24\textwidth]{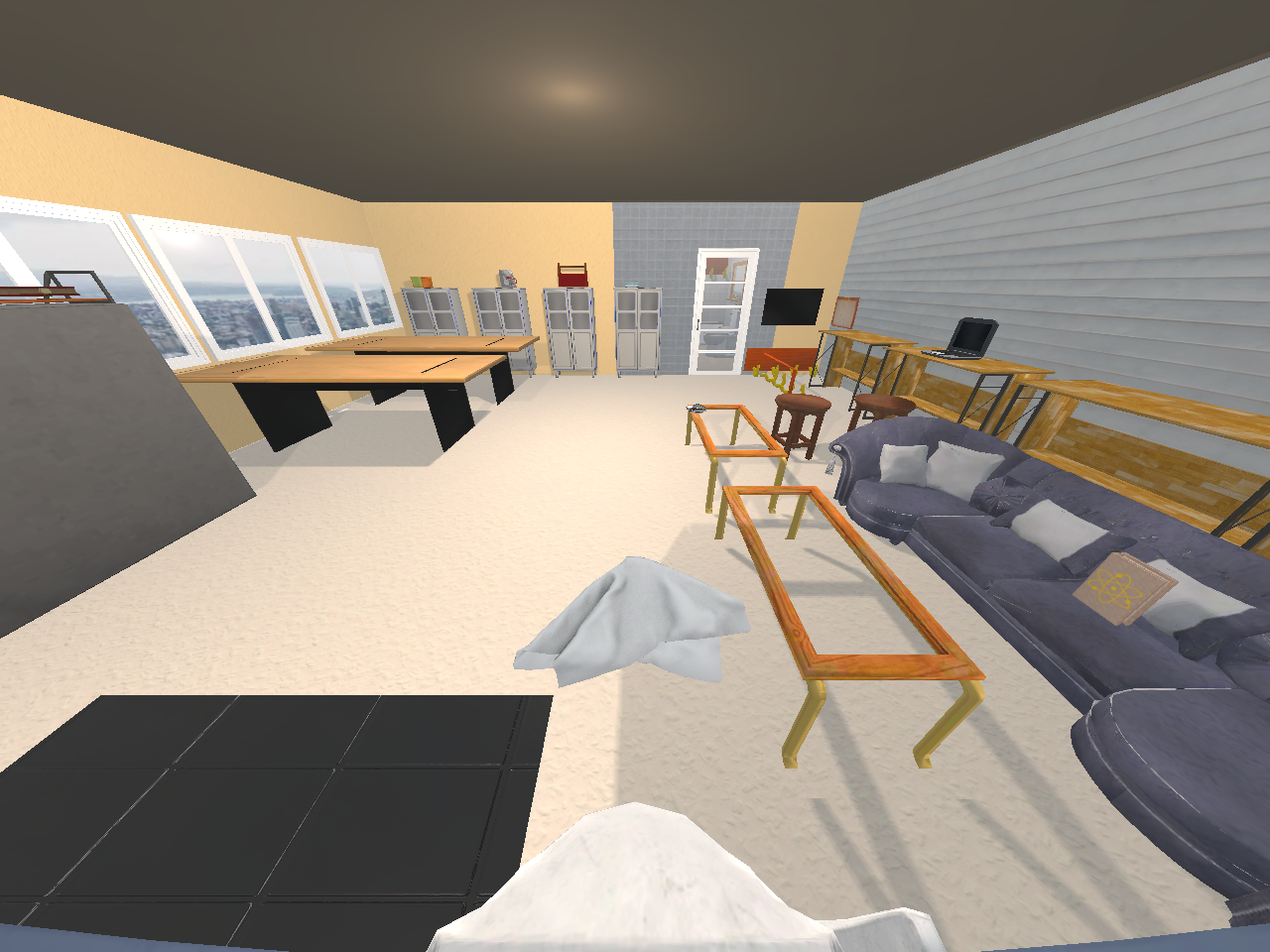}
\end{minipage}
\end{tcolorbox}
\caption{Sample screenshots from a task execution. The robot begins on the second floor, takes the elevator to the first floor to retrieve a cloth, and then returns to the sofa. \emph{Task query: I want to clean my sofa. Go get a cloth from the first floor and come back near the sofa.}}
\label{fig:execution}
\end{figure*}

In single-floor tasks, the agent is required to locate an object (e.g. basketball, laptop) and deliver it to another room. In two-floor tasks, the agent is asked to first get a cloth from the first floor and deliver it to the second floor. The example is shown in Fig.~\ref{fig:execution}. In the four-story setting, the agent starts on the first floor, collects an orange toolbox on the third floor, and delivers it to the fourth floor.

\section{Skills in MANSION}

\subsection{Skill library expansion in MANSION}

To better support the baseline algorithms in MANSION, we extend the original AI2-THOR skill library with three essential atomic skills required for multi-floor, long-horizon tasks: \emph{CallElevator}, \emph{UseElevator}, and \emph{TakeStairs}. The detailed descriptions of these skills can be found in the following skill cards.

\begin{ToolCard}{Skill: UseElevator}
    \ToolKey{Description}  
    Selects the target floor and performs a floor transition using the elevator if a valid floor number is provided. State information transition will be processed internally.
    
    \ToolKey{Prerequisite}  
    \begin{itemize}
        \item The robot must already be positioned at the elevator entrance. 
        \item CallElevator skill is used.
    \end{itemize}

    \ToolKey{Input} \texttt{target\_floor}

    \ToolKey{Internal process}  
    The parameter \texttt{target\_floor} is converted into a floor offset: \texttt{floor\_delta} = \texttt{target\_floor} - \texttt{current\_floor}, which is passed to \texttt{ThorGym.update\_floor(floor\_delta)}.  
    
    \texttt{update\_floor(floor\_delta)} internally:
    \begin{enumerate}
        \item Removes the currently held object from the current-floor JSON and inserts it into the target-floor JSON.
        \item Calls \texttt{controller.reset(scene=\dots)} to reload the target floor’s scene.
        \item Reconstructs room landmarks and the room graph.
        \item Teleports the robot to a standardized starting pose outside the elevator on the new floor, oriented outward from the elevator.
    \end{enumerate}
\label{tool:useelevator}
\end{ToolCard}

\begin{ToolCard}{Skill: CallElevator}
    \ToolKey{Description}  
    Call the elevator to the robot’s current floor and open the elevator door.

    \ToolKey{Prerequisite}  
    \begin{itemize}
        \item The robot must be located near the elevator entrance.
    \end{itemize}

    \ToolKey{Input} None

    \ToolKey{Internal process}  
    Invokes the environment’s elevator-calling mechanism:
    \begin{enumerate}
        \item Sends a call request to bring the elevator to \texttt{current\_floor}.
        \item Open the elevator door for the next step operation.
    \end{enumerate}

\label{tool:callelevator}
\end{ToolCard}

\begin{ToolCard}{Skill: TakeStairs}
    \ToolKey{Description}  
    Move the robot between adjacent floors using the staircase. This skill serves as an alternative to elevator-based floor transitions. The VLM determines whether the robot should go Up or Down based on the stair-view image.

    \ToolKey{Prerequisite}  
    \begin{itemize}
        \item The robot must be positioned at the staircase entrance.
        \item The direction must correspond to an available adjacent floor.
    \end{itemize}

    \ToolKey{Input} \texttt{direction} $\in \{\text{Up}, \text{Down}\}$

    \ToolKey{Internal process}  
    The parameter \texttt{direction} is mapped to a floor offset: \texttt{floor\_delta} =
       $ \begin{cases}
            +1 & \text{if direction = Up},\\
            -1 & \text{if direction = Down}.
        \end{cases}$ This \texttt{floor\_delta} is passed to \texttt{ThorGym.update\_floor(floor\_delta)}.

\label{tool:takestairs}
\end{ToolCard}

\subsection{Progress Score}
We decompose task completion into two components: correct object retrieval and successful navigation, as described in Section 4.2. This separation reflects a key limitation of current VLMs: they struggle to reliably identify and retrieve small objects, even though they possess a stronger, more global understanding of room layout and spatial context. Therefore, in addition to reporting overall success rates, we also evaluate performance using a progress score that captures partial task completion.

\subsection{Task details}

We now provide the detailed prompts that we used in the Table~\ref{tab:task_settings_prompts_fixed}.

\begin{table}[t]
    \centering
    \caption{Task Settings and Prompts}
    \label{tab:task_settings_prompts_fixed}
    \begin{tabularx}{\columnwidth}{%
        >{\centering\arraybackslash}p{0.25\columnwidth}  
        >{\raggedright\arraybackslash}X                  
    }
        \toprule
        \textbf{Environment} & \textbf{Prompt} \\
        \midrule
        Single-floor apartment &
        Find a box on the bed and bring it to the bathroom. \\
        \midrule
        Double-floor office &
        Find a cellphone on a blue couch on the first floor and bring it
        to the round table on the second floor. \\
        \midrule
        Four-floor office &
        Go to the third floor, find a laptop on the desk in the meeting
        room on the third floor, and take it to the restroom on the
        fourth floor. \\
        \bottomrule
    \end{tabularx}
\end{table}

Some sample test environments that we used can be found in the following Fig.~\ref{fig:onefloor}--\ref{fig:fourfloor}.

\begin{figure}[h]
    \centering
    \includegraphics[width=\linewidth]{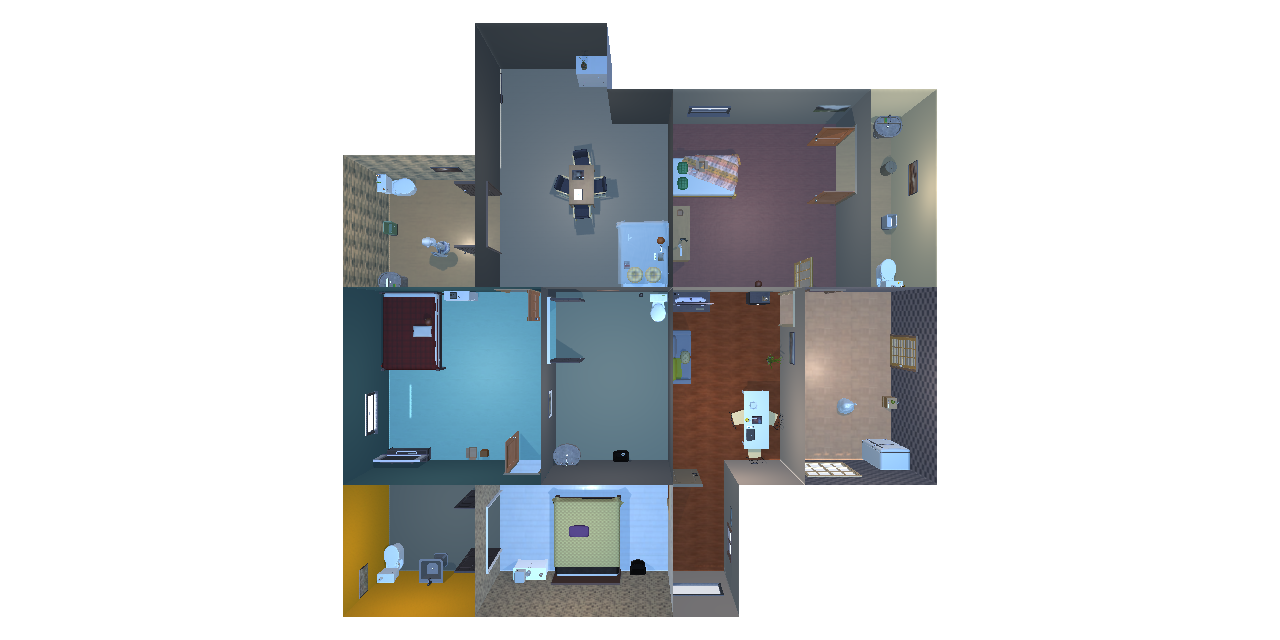}
    \caption{Single-floor apartment layout.}
    \label{fig:onefloor}
\end{figure}

\begin{figure}[h]
    \centering
   \includegraphics[
    trim={0.05\linewidth, 0, 0, 0},
    clip,
    width=\linewidth
]{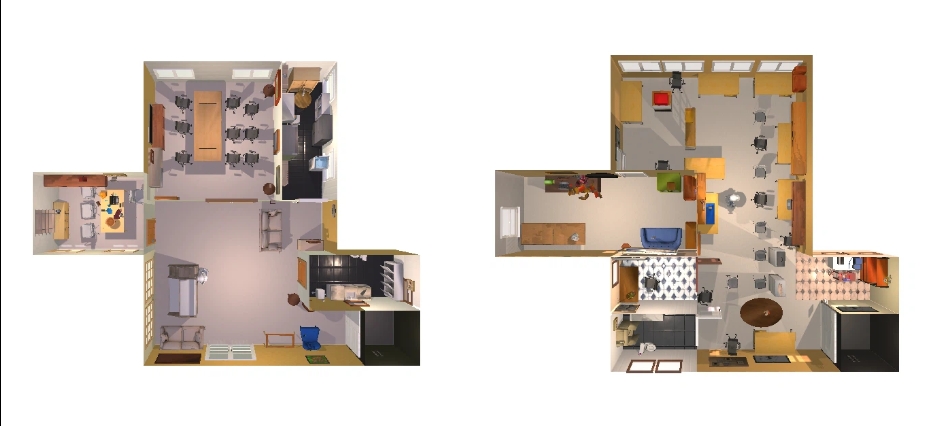}
    \caption{Two-floor office layout.}
    \label{fig:twofloor}
\end{figure}

\begin{figure}[h]
    \centering
    \includegraphics[width=\linewidth]{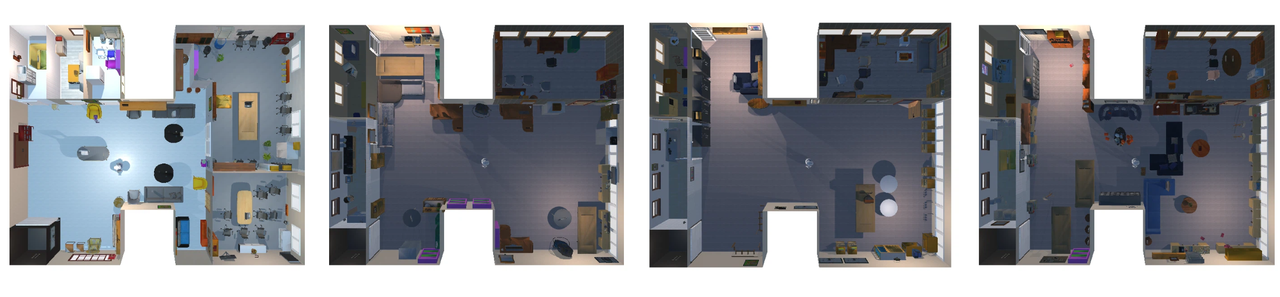}
    \caption{Four-floor building layout.}
    \label{fig:fourfloor}
\end{figure}

\subsection{Algorithms implementation details}
In integrating the embodied algorithms into MANSION, we introduce several key adaptations to better reflect the robot’s actual capabilities within the simulated environment.

\begin{figure}[h]
    \centering
    \includegraphics[width=\linewidth]{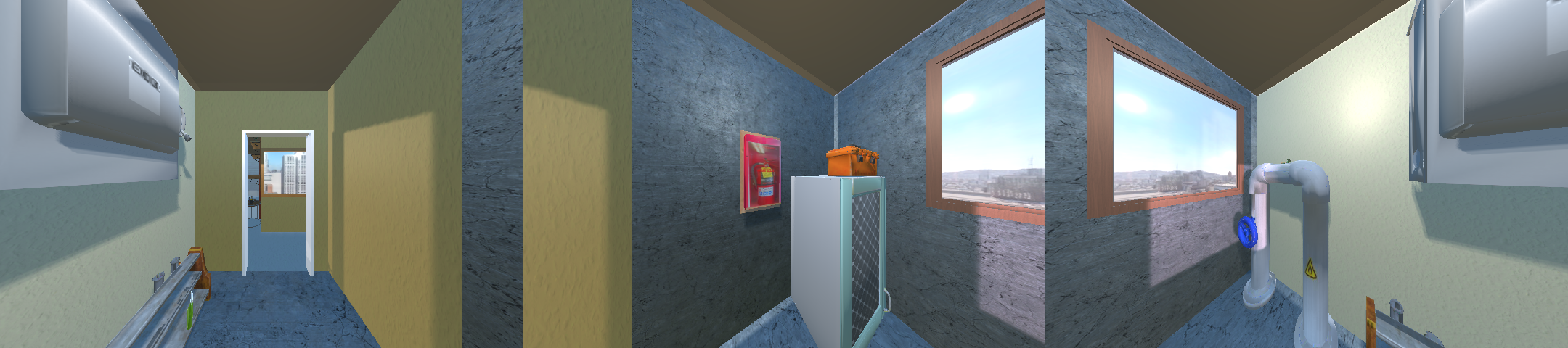}
    \caption{Panorama view of the service shaft room.}
    \label{fig:service_shaft_panorama}
\end{figure}

\textbf{GoToLandmark: } The success of embodied navigation algorithms depends heavily on the VLM’s ability to obtain reliable visual observations of different rooms. The robot can only plan a route to the correct destination if the VLM correctly identifies the room type. However, in the original BUMBLE implementation, each room is represented by a single image. When that image happens to capture a featureless or uninformative part of the room, the robot’s failure rate increases significantly. To address this, we provide panorama views for each room, giving the robot a more complete and informative visual representation of the environment and improving its ability to recognize the correct room. An example can be found in Fig.~\ref{fig:service_shaft_panorama}. To balance image granularity with the input constraints of VLMs, each room’s panorama is constructed by concatenating three images captured at yaw angles of 0°, 120°, and 240°.

\textbf{UseElevator: } The agent is informed of its current floor and given a visual observation that includes the elevator button panel. It must identify the valid floor numbers from the panel and select the target floor it intends to reach.

\textbf{TakeStairs: } The agent is told of its floor and provided with the corresponding visual observation. To prevent invalid decisions, such as attempting to go downstairs from the first floor, we overlay valid directional arrows onto the visual input, ensuring the agent is guided toward only feasible movement options.

\subsection{Failure Case Analysis}
\label{sec:fail}
In this subsection, we analyze several representative task failure cases and their underlying causes.

\begin{figure}[h]
    \centering
    \includegraphics[width=\linewidth]{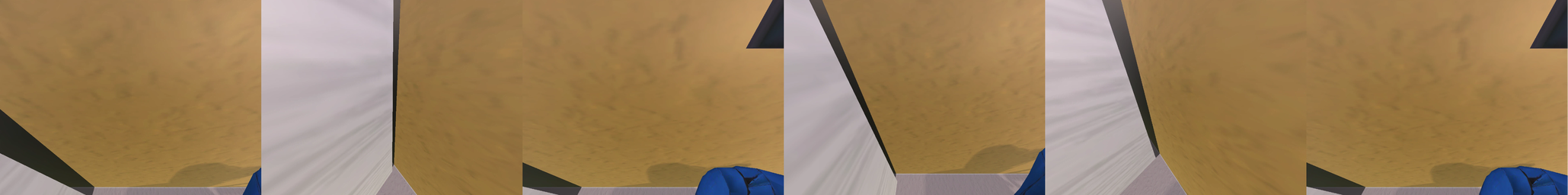}
    \caption{Failure case 1}
    \label{fig:fail_two_floor}
\end{figure}

\noindent\textbf{Failure Case Analysis 1: Two-floor task.}
A typical failure pattern is as follows: the agent navigates into a corner and, even after attempting to backtrack and rotate, still cannot escape from the corner, eventually exhausting the step budget and failing the task. See in Fig.~\ref{fig:fail_two_floor}

\begin{figure}[h]
    \centering
    \includegraphics[width=\linewidth]{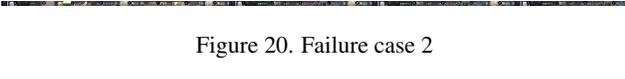}
    \caption{Failure case 2}
    \label{fig:fail_four_floor}
\end{figure}

\noindent\textbf{Failure Case Analysis 2: Four-floor task.}
The most prominent issue is that \texttt{goto\_landmark} needs to stitch all landmarks into a single long image as input to the VLM. However, in the four-floor building, there are too many landmarks, so the stitched image must be heavily downsampled when resized to the VLM input resolution, causing severe information loss and making it difficult for \texttt{goto\_landmark} to function effectively. See in Fig.~\ref{fig:fail_four_floor}

\section{Object Placement}
\label{sec:obj_placement}

For complex rooms, the key challenge lies not only in accommodating a larger number
of objects, but also in preserving regular global distribution under dense and
repeated furniture patterns. A purely instance-level object placement algorithm tends to
over-emphasize local relations, which can overfill one part of a large room
while leaving other feasible regions unused. We shift our planning focus from individual objects to structured groups before solving the geometry.

We require the LLM to output object-level constraints for each item, including three types of constraints: global placement constraints (e.g., \texttt{edge}, \texttt{middle}, or \texttt{unconstrained}), structural constraints (e.g.,
\texttt{single}, \texttt{matrix}, or \texttt{paired}), and optional relative position constraints (e.g., \texttt{near}, \texttt{far}, etc.). The \texttt{matrix} primitive compactly represents repeated rows such as desk rows
or bookshelf blocks, while \texttt{paired} expresses one-to-one accessory
relations such as desk--chair pairs. This representation reduces the burden on
the LLM, since in large spaces such as classrooms, libraries, or offices, it no
longer needs to output dozens of nearly identical instance-level constraints.
We then normalize these constraints into groups $G=(a,M)$, where $a$
denotes the anchor object and $M$ denotes the member set. The anchor carries the
global spatial role of the group, while the remaining members are placed in the
anchor's local frame.

\begin{algorithm}[t]
  \caption{Priority-aware group-based object placement}
  \label{alg:object-placement}
  \begin{algorithmic}[1]
    \Require Room polygon $\Omega$; normalized object groups $\mathcal{G}$; placement constraints $\mathcal{C}$
    \Ensure Placement set $\mathcal{P}$
    \State Sort $\mathcal{G}$ by the constraints of $a(G)$:
    \Statex \hspace{1.5em} $\texttt{edge+matrix} \succ \texttt{edge} \succ \texttt{matrix} \succ \texttt{middle} \succ \texttt{free}$
    \State $\mathcal{P} \gets \emptyset$
    \For{\textbf{each} group $G$ in $\mathcal{G}$}
      \If{$a(G)$ has a \texttt{matrix} constraint}
        \State $(r, c) \gets$ requested matrix size of $a(G)$
        \State $\mathcal{Q} \gets \emptyset$
        \While{$r \ge 1$ and $c \ge 1$ and $\mathcal{Q} = \emptyset$}
          \State $\hat{o} \gets \textsc{BuildMacroObject}(G, r, c)$
          \State $\mathcal{Q} \gets \textsc{FindFeasiblePlacement}(\hat{o}, \Omega, \mathcal{P}, \mathcal{C})$
          \If{$\mathcal{Q} = \emptyset$}
            \State $(r, c) \gets \textsc{DowngradeMatrix}(r, c)$
          \EndIf
        \EndWhile
        \If{$\mathcal{Q} \neq \emptyset$}
          \State $\mathcal{P} \gets \mathcal{P} \cup \mathcal{Q}$
        \EndIf
      \Else
        \For{\textbf{each} object $o$ in $G$}
          \State $\mathcal{Q} \gets \textsc{FindFeasiblePlacement}(o, \Omega, \mathcal{P}, \mathcal{C})$
          \If{$\mathcal{Q} \neq \emptyset$}
            \State $\mathcal{P} \gets \mathcal{P} \cup \mathcal{Q}$
          \EndIf
        \EndFor
      \EndIf
    \EndFor
    \State \Return $\mathcal{P}$
  \end{algorithmic}
\end{algorithm}

We present our object placement algorithm in Algorithm~\ref{alg:object-placement}. Our algorithm follows a priority-aware constructive
search. Groups are sorted according to the constraints of their anchor object,
yielding a strict priority order. Groups that are both wall-dependent and highly
structured are processed first, since they occupy the most constrained regions
of the room and strongly affect later circulation. For a matrix group, the
solver first places the whole pattern as a macro object; if no feasible
placement is found, it progressively downgrades the matrix size and retries. For
a non-matrix group, objects are processed sequentially within the group,
starting from the anchor object. For each object, the solver samples candidate
positions and filters them by hard constraints including collision checking,
constraint consistency, and incremental reachability. Objects that do not admit
a feasible placement are discarded, while the solver continues with the
remaining objects.

\begin{figure}[h]
  \centering
  \includegraphics[width=\linewidth]{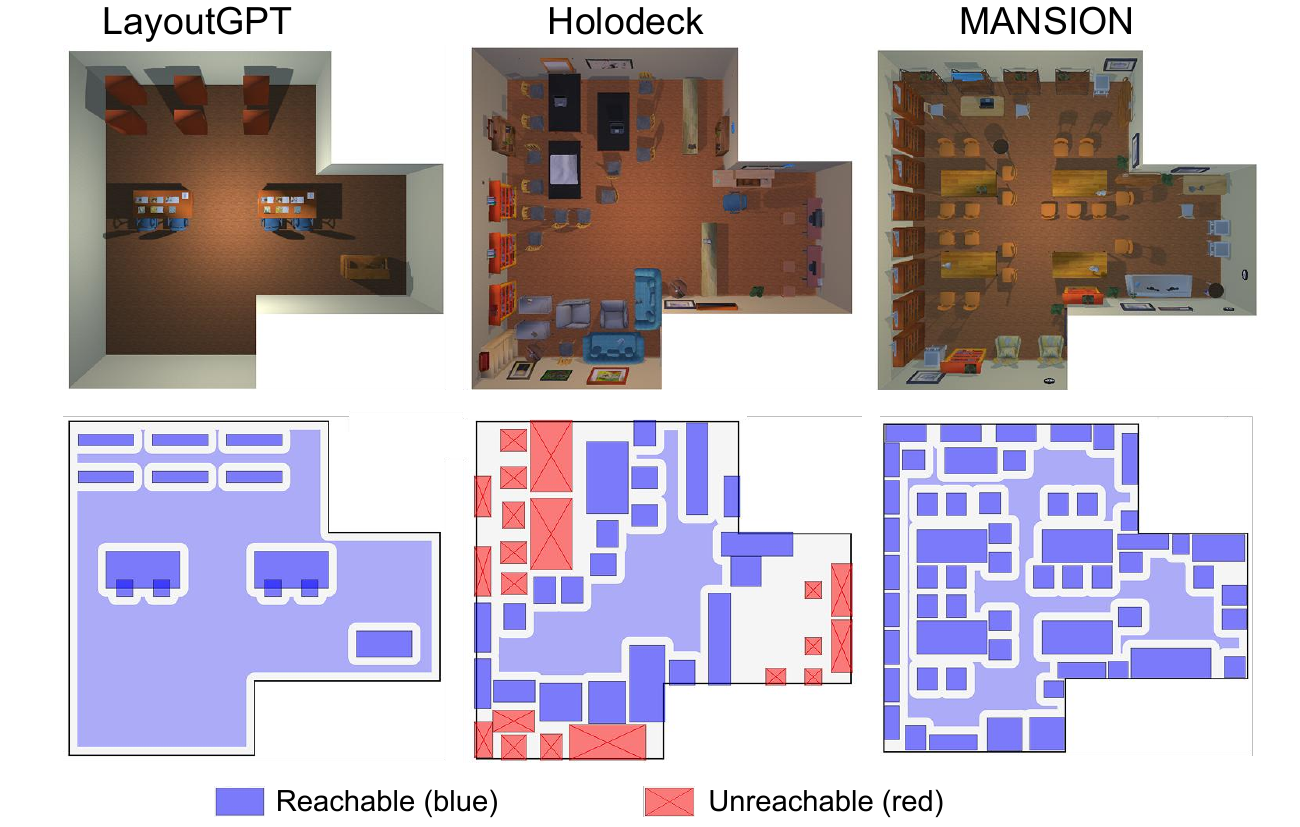}
  \caption{Reachability visualization in a library scene.}
  \label{fig:library-reachability}
\end{figure}

Reachability is evaluated on the remaining free space when searching for
feasible positions, ensuring that the room entrance remains connected to the
required circulation areas and to accessible interaction zones around the placed
objects. Candidates that block passages or destroy walkable structures are
discarded immediately. As shown in the Fig.~\ref{fig:library-reachability}, our method maintains full reachability while preserving a high object placement count.

\section{User Study}
\label{sec:user_study}
\begin{figure}[h]
    \centering
    
    \begin{subfigure}{\linewidth}
        \centering
        \includegraphics[width=\linewidth]{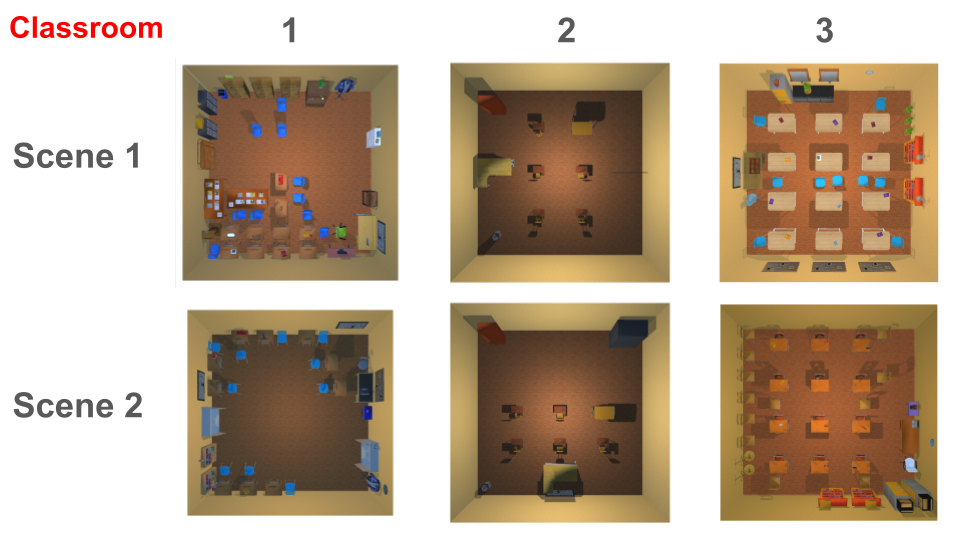}
        \\[1ex] 
        \includegraphics[width=\linewidth]{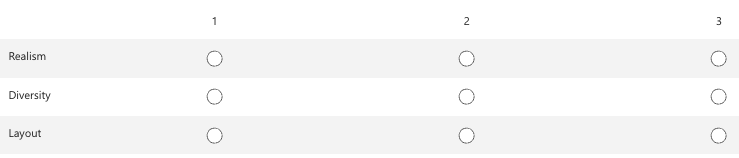}
    \end{subfigure}
    \vspace{-10pt}
    \caption{Sample from the user study for the classroom scene category.}
    \label{fig:survey}
\end{figure}

To understand how real-person users perceive our generated scenes compared to other methods, we conducted a comprehensive user study with 52 participants from different backgrounds. The full list of participating institutions can be found in Section~\ref{sec:acknowledge}, and we thank them again for their input. For each scene type, we randomly sample two cases from the 10 generated results with the same prompt for subjective evaluation. In each scene setting, participants are presented with one set of images from the three methods and are asked to select the best method in terms of realism, diversity, and overall layout quality. To prevent bias, we made sure that the recruited participants had no prior experience or exposure to 3D scene generation. Furthermore, we kept the names of the corresponding algorithms hidden from them throughout the survey. A sample of the user study image and form can be seen in Fig.~\ref{fig:survey}.

We also provide the metric instructions that we used in the survey to guide the users to rank the different methods below.

\begin{tcolorbox}[colback=gray!15, colframe=gray!15, arc=3mm, auto outer arc]
\textbf{\large Instructions}
\begin{itemize}
    \item Realism: How realistic and plausible the scene is (object choices and placements make sense in the given setting; no obvious weird placements or impossible arrangements). Which one is most realistic to real life?
    
    \item Diversity: How different and varied the generated scenes are across different generated scenes of the SAME method (variety in object selection, arrangement, and overall design, while still matching the scenario)

    \item Layout: How well-organized and functional the spatial arrangement is (clear structure, reasonable spacing, good flow, and sensible grouping of objects).
\end{itemize}
\end{tcolorbox}

\section{Prompt Templates}
\label{app:prompts}

We present prompt templates for three representative modules:
(i) whole-building program planning,
(ii) single-floor topology (bubble graph) generation, and
(iii) LLM-guided seed box selection for cutting.
For readability, these templates preserve the core task definition, input fields, output schema, and major constraints, while abstracting away some implementation-level details. The templates are shown in Fig.~\ref{fig:prompt_whole_building}--\ref{fig:prompt_seed_cutting}.

\begin{figure*}[t]
    \centering
    \begin{ToolCard}{Prompt Template: Whole Building Planning}
        \small 
        \ToolKey{Module:} \texttt{building\_program\_planner}
        
        \ToolKey{System Instruction:}
        You are an experienced architect asked to program a complete multi-floor building using the first-floor plan as reference.

        \textbf{Inputs:}
        \begin{itemize}[noitemsep, topsep=0pt, leftmargin=*]
            \item \textbf{Visual:} \textit{[Image Input: Base64 rasterized boundary]}
            \item \textbf{Geometry:} First-floor boundary JSON
            \item \textbf{Params:} Floors: \texttt{\{floors\}}, \texttt{\{area\_note\}}
            \item \textbf{Requirement:} \texttt{\{user\_part\}}
        \end{itemize}

        \textbf{Core Placement Preferences:}
        \begin{itemize}[noitemsep, topsep=0pt, leftmargin=*]
            \item Place stairs/elevators only in corners that are bounded by two exterior walls; pick the corner whose surrounding leftover space is smallest to keep large continuous areas intact.
            \item Make use of tight/awkward leftover pockets.
            \item Quantize core boxes to integer coordinates and size exactly \texttt{\{core\_area\}}. Coordinates must be non-negative.
        \end{itemize}

        \textbf{What to Do:}
        \begin{enumerate}[noitemsep, topsep=0pt, leftmargin=*]
            \item Analyze the outline and area and reason about a practical building function.
            \item Decide the vertical connectivity method: stair $|$ elevator $|$ stair\_and\_elevator, consistent with the rule.
            \item Choose locations for vertical cores; each stair/elevator occupies an axis-aligned \texttt{\{core\_area\}} bbox within the floor polygon. Output as x=[x1,x2], y=[y1,y2] with integer coordinates.
            \item For each floor, produce a list ``rooms" with ID-only entries, area estimates, and material specifications. \textbf{Do NOT include a ``type" field.} The \textbf{first room} in the list MUST be the circulation hub of the floor (traffic core): choose logically based on building type. Each room must include \texttt{floor\_material} and \texttt{wall\_material} fields with descriptive text.
            \item Room sums must not exceed the gross floor area; keep 12--25\% as circulation/core reserve unless justified.
            \item Ensure totals are reasonable; indicate whether plans fit within GFA.
            \item If a floor layout should be exactly the same as another floor, use a shorthand: specify only \texttt{\{"index": k, ``copy": j\}}.
        \end{enumerate}
        
        \ToolKey{Output Schema (Strict JSON):}
        {
        \scriptsize 
\begin{verbatim}
{
  "reasoning": "<Brief explanation of program and area logic>",
  "vertical_connectivity": {
    "method": "stair | elevator | stair_and_elevator",
    "cores": [ {"type": "stair", "x": [<int>, <int>], "y": [<int>, <int>]} ],
    "justification": "<Why this choice fits rules and requirements>"
  },
  "floors": [
    {
      "index": <int>,
      "requirement": "<Natural language requirement for this floor>",
      "gross_floor_area": <float>,
      "rooms": [
        {
          "id": "hub_room", "area_estimate": <float>, 
          "floor_material": "<Description (e.g., warm oak hardwood, matte)>", 
          "wall_material": "<Description (e.g., soft beige drywall, smooth)>", 
          "notes": "circulation hub (put FIRST)"
        },
        {
          "id": "<other_room_id>", "area_estimate": <float>,
          "floor_material": "<Description>", "wall_material": "<Description>",
          "notes": "<Optional>"
        }
      ],
      "area_summary": {
        "sum_rooms": <float>, "reserve_ratio": <float>,
        "fits_within_gfa": <boolean>, "notes": "<Optional notes>"
      }
    }
  ]
}
\end{verbatim}
        }

        \textbf{Floor Layout Context:} \texttt{\{layout\_json\}}
    \end{ToolCard}
    \caption{Prompt template for whole-building program planning.}
    \label{fig:prompt_whole_building}
\end{figure*}

\begin{figure*}[t]
    \centering
    \begin{ToolCard}{Prompt Template: Single-Floor Topology Generation}
        \small
        \ToolKey{Module:} \texttt{topology\_bubble\_planner}

        \ToolKey{System Instruction:}
        You are an experienced architect designing the abstract topological connectivity of a single floor.

        \ToolKey{Inputs:}
        \begin{itemize}[noitemsep, topsep=0pt, leftmargin=*]
            \item Overall program reasoning: \texttt{\{reasoning\}}
            \item Floor context: Floor index \texttt{\{idx\}}, Gross floor area \texttt{\{gfa\}} $m^2$, Floor requirement \texttt{\{requirement\}}
            \item Floor polygon JSON (main space after cores removed): \texttt{\{layout\_json\}}
            \item Vertical cores: \texttt{\{vtext\}}
            \item Floor hints from program: \texttt{\{rooms\_json\}} (first item is the suggested circulation hub)
            \item Material selection guidance: \texttt{\{material\_hints\_text\}}
        \end{itemize}

        \ToolKey{Your Task:}
        \begin{itemize}[noitemsep, topsep=0pt, leftmargin=*]
            \item Derive a minimal useful set of rooms/spaces for this floor based on the requirement and rooms list, and assign each node an estimated area ($m^2$); capture only abstract connectivity, not geometry.
            \item Include all elevator/stair connectors indicated by the floor layout as fixed nodes for this floor; do not omit them.
            \item Treat the provided rooms list as hints (the first item is the suggested circulation hub), but re-evaluate which space should serve as \texttt{main} if one space clearly dominates the floor.
        \end{itemize}

        \ToolKey{Output Requirements:}
        \begin{itemize}[noitemsep, topsep=0pt, leftmargin=*]
            \item Return \textbf{exactly one JSON object} with two top-level fields: \texttt{nodes} and \texttt{edges}, with no extra explanatory text.
            \item \textbf{Required node fields:} \texttt{id}, \texttt{type}, \texttt{area}, \texttt{floor\_material}, \texttt{wall\_material}, \texttt{open\_relation}.
            \item \texttt{Node types} allowed: \texttt{main}, \texttt{Entities}, \texttt{area}, \texttt{elevator}, \texttt{stair}.
            \item \texttt{Edge kinds} allowed: \texttt{access}, \texttt{adjacent}.
            \item Every node must include \texttt{floor\_material} and \texttt{wall\_material} using descriptive text.
            \item \texttt{open\_relation} must be either ``open'' or ``door''. For \texttt{main}, use ``open''; for \texttt{elevator} and \texttt{stair}, use ``door''.
        \end{itemize}

        \ToolKey{Design Preferences:}
        \texttt{\{node hierarchy \& branching\}}, \texttt{\{area vs.\ Entities selection\}}, \texttt{\{open\_relation assignment\}}
        
        \ToolKey{Output Schema:}
        {
        \scriptsize
\begin{verbatim}
{
  "nodes": [
    {"id": "lobby", "type": "main", "area": 80.0,
     "floor_material": "warm oak hardwood, matte",
     "wall_material": "soft beige drywall, smooth", "open_relation": "open"},
    {"id": "office_zone", "type": "area", "area": 30.0,
     "floor_material": "warm oak hardwood, matte",
     "wall_material": "soft beige drywall, smooth", "open_relation": "door"},
    {"id": "room_1", "type": "Entities", "area": 15.0,
     "floor_material": "carpet, neutral gray",
     "wall_material": "painted drywall, white", "open_relation": "door"}
  ],
  "edges": [
    {"source": "lobby", "target": "office_zone", "kind": "adjacent"},
    {"source": "office_zone", "target": "room_1", "kind": "adjacent"}
  ]
}
\end{verbatim}
        }
    \end{ToolCard}
    \caption{Prompt template for single-floor topology generation.}
    \label{fig:prompt_topology}
\end{figure*}

\begin{figure*}[t]
    \centering
    \begin{ToolCard}{Prompt Template: Hierarchical Seed Planning}
        \small
        \ToolKey{Module:} \texttt{seed\_guidance}

        \ToolKey{System Instruction:}
        You are a floor plan seed planning assistant. Given the floor topology and a preview image containing the full floor outline with stairs/elevators, your task is to plan an axis-aligned rectangular bounding box for each target room within the current parent room, expressed as \texttt{x=[xmin,xmax], y=[ymin,ymax]} to indicate each room's approximate position and extent.
        \ToolKey{Context provided:}
        \begin{itemize}[noitemsep, topsep=0pt, leftmargin=*]
            \item Target room list: \texttt{\{target\_ids\_text\}}
            \item Parent room ID: \texttt{\{parent\_id\}}, Type: \texttt{\{parent\_type\}}, Area: \texttt{\{parent\_area\}} $m^2$
            \item Adjacent room IDs (already placed): \texttt{\{neighbor\_ids\_text\}}
            \item Project requirement: \texttt{\{requirement\}}
            \item Topology JSON and area hints per room
            \item \{special\_instruction\}
        \end{itemize}
        \textbf{Coordinate convention:} $x$ increases to the right, $y$ increases upward; the parent room's approximate coordinate range is $x \in [\texttt{\{minx\}}, \texttt{\{maxx\}}]$, $y \in [\texttt{\{miny\}}, \texttt{\{maxy\}}]$; your output should use values within this coordinate interval.

        \textbf{Output format:} The output must be a JSON array where each element contains:
        \begin{itemize}[noitemsep, topsep=0pt, leftmargin=1.5em]
            \item \texttt{room\_id}: string, the room ID (must be chosen from the candidate list only)
            \item \texttt{x}: array of length 2 \texttt{[xmin,xmax]}, the bounding box in the $x$ direction, requires $\texttt{xmin} \le \texttt{xmax}$
            \item \texttt{y}: array of length 2 \texttt{[ymin,ymax]}, the bounding box in the $y$ direction, requires $\texttt{ymin} \le \texttt{ymax}$
            \item \texttt{area}: float, the approximate fraction of the parent room's area this room occupies (0.0--1.0, optional, used as a downstream hint)
            \item \texttt{reason}: a brief one-sentence explanation of why you placed this room at this position
        \end{itemize}

        \ToolKey{Output Schema (Strict JSON):}
        {
        \scriptsize
\begin{verbatim}
[
  {"room_id": "lobby", "x": [0.0, 8.0], "y": [0.0, 6.0], "area": 0.45,
   "reason": "Main lobby occupies the central area, enclosing the stair core."},
  {"room_id": "office_1", "x": [8.0, 12.0], "y": [0.0, 5.0], "area": 0.2,
   "reason": "Office placed at the east wing, away from the stair core."}
]
\end{verbatim}
        }

        Please refer to the floor outline and existing stair/elevator positions in the preview image to provide a reasonable bounding box allocation. Note: bounding boxes should be placed as much as possible within the parent room outline; avoid large-area overlaps.
    \end{ToolCard}
    \caption{Prompt template for LLM-guided seed box planning for cutting.}
    \label{fig:prompt_seed_cutting}
\end{figure*}

\end{document}